\documentclass[10pt,twocolumn,letterpaper]{article}

\usepackage[pagenumbers]{cvpr} %

\definecolor{cvprblue}{rgb}{0.21,0.49,0.74}
\usepackage[pagebackref,breaklinks,colorlinks,allcolors=cvprblue]{hyperref}

\def\paperID{11516} %
\def\confName{CVPR}
\def\confYear{2026}

\title{FedPromo: Federated Lightweight Proxy Models At The Edge \\Bring New Domains To Foundation Models}

\author{Matteo Caligiuri, Francesco Barbato, Donald Shenaj, Umberto Michieli, Pietro Zanuttigh\\
Department of Information Engineering\\
University of Padova, 
Padova, Italy \\
{\tt\small \{caligiurim, barbatofra, shenajdona, michieli, zanuttigh\}@dei.unipd.it}
}

\usepackage{makecell}
\usepackage{xfrac}
\usepackage[table]{xcolor} %
\usepackage{fdsymbol}
\usepackage{pifont}
\usepackage{fontawesome}
\usepackage{algorithm}
\usepackage{algorithmic}
\usepackage{multirow}
\usepackage{arydshln}
\usepackage{placeins}

\graphicspath{{figures/}}

\newcommand{\cmark}{\ding{51}}%
\newcommand{\xmark}{\ding{55}}%

\newcommand{\sname}{CA-FKT}%

\begin{document}

\maketitle

\begin{abstract}
Federated Learning (FL) is an established paradigm for training deep learning models on decentralized data. However, as the size of the models grows, conventional FL approaches often require significant computational resources on client devices, which may not be feasible. 
We introduce FedPromo, a novel framework that enables efficient adaptation of large-scale foundation models stored on a central server to new domains encountered only by remote clients. 
Instead of directly training the large model on client devices, FedPromo optimizes lightweight proxy models via FL, significantly reducing computational overhead while maintaining privacy.
Our method follows a two-stage process: first, server-side knowledge distillation aligns the representations of a large-scale foundation model (\eg, a transformer) with those of a compact counterpart (\eg, a CNN). Then, the compact model encoder is deployed to client devices, where trainable classifiers are learned locally. These classifiers are subsequently aggregated and seamlessly transferred back to the foundation model, facilitating personalized adaptation without requiring direct access to user data.
Through novel regularization strategies, our framework enables decentralized multi-domain learning, balancing performance, privacy, and resource efficiency. 
Extensive experiments on five image classification benchmarks demonstrate that FedPromo outperforms existing methods while assuming limited-resource clients.
\end{abstract}

\renewcommand{\thefootnote}{\roman{footnote}}
\footnotetext[0]{\raggedright Code and dataset splits are available at: \mbox{\url{https://github.com/LTTM/FedPromo}}}
\renewcommand{\thefootnote}{\arabic{footnote}}

\section{Introduction} \label{sec:intro}
\begin{figure}[t]
    \centering
    \includegraphics[width=.9\linewidth]{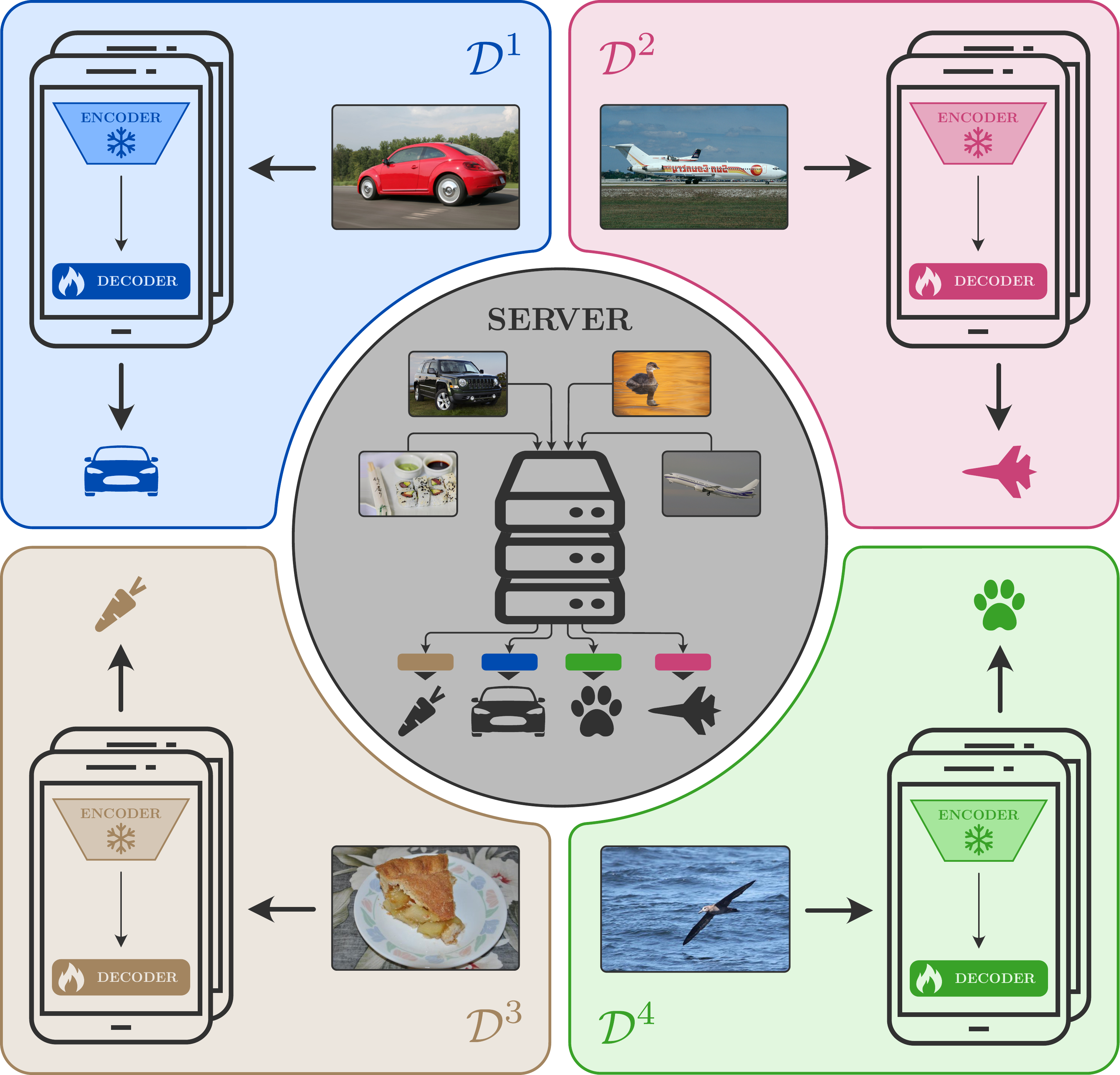}
    \caption{FedPromo enables large-scale model adaptation via Federated Learning through compact proxy models. By combining cross-architectural distillation with specialized training strategies, it achieves private, decentralized, multi-domain adaptation to data on resource-constrained clients.}
    \label{fig:graphabs}
\end{figure}

In recent years, %
remarkable progresses in visual understanding tasks %
have been achieved thanks to high-performance foundation models (FM) based on complex transformer architectures~\cite{radford2021learningtransferablevisualmodels,sam,oquab2023dinov2,liu2023visualinstructiontuning}. However, deploying such models on edge devices remains challenging due to their substantial computational demands~\cite{10.1145/3505244,10.1145/3530811}.
Despite improvements in personal devices such as smartphones, smart home assistants, and connected vehicles, they still cannot efficiently run large-scale foundation models~\cite{10.1145/3530811}. This creates a critical gap between  potential performances and practical deployment in contexts where privacy and resource limitations are crucial. In fact, real-world applications often generate valuable user data on edge devices~\cite{sethi2017internet}, but transferring it to centralized servers is often infeasible due to privacy issues, security concerns, and high communication costs.
This raises a key question: is it possible to transfer fine-grained private client information to a server-side foundation model %
which can not access the private data on the devices and with minimal resources at the client side?

Federated Learning (FL)~\cite{Zhang21survey,wen2023survey,shenaj2023federated}, offers a natural solution to learning in decentralized settings that allows local model updates without exposing raw user data to a central server. However, conventional %
approaches assume that clients can locally train and execute the entire model~\cite{mcmahan2017communication}, which becomes impractical for large-scale models that exceed the computational constraints of edge devices.

To address this limitation, we introduce the Cross-Architecture Federated Knowledge Transfer (CA-FKT) paradigm (Sec.~\ref{sec:formulation}). It enables decentralized training of lightweight proxy models for fine-grained datasets, integrating their knowledge into a large-scale foundation model that cannot be deployed on clients. Then, we propose a novel framework, FedPromo (see Fig.~\ref{fig:graphabs}) to tackle it.
Our approach works in two stages, combining the strengths of cross-architectural Knowledge Distillation (KD) and FL:
\begin{enumerate}
    \item \textbf{Server-side Cross-Architectural KD}:
    Before deploying proxy models to clients, we align the feature space of their lightweight CNN encoder (student) to that of a large-scale transformer-based model (teacher).
    This %
    ensures that the small computationally efficient model can 
    approximate the embeddings of the larger model %
    \cite{barbato2024cross,hao2023oneforallbridgegapheterogeneous}.
    \item \textbf{Federated Training of Proxy Models}: Proxy encoders are deployed to client devices and kept frozen to ensure consistency and minimize computational cost~\cite{sidahmed2021efficientprivatefederatedlearning,Chen_Zhang_Krompass_Gu_Tresp_2024,9546506}.
    The local optimization concerns a single classifier layer which, after training, is aggregated on the server and seamlessly transferred to the foundation model, enabling multi-domain adaptation without direct data access.
\end{enumerate}

\noindent By freezing the encoder and training only the classifier, we achieve two key objectives:
(1) \textit{Personalization}: clients adapt their classifiers to local data domains while benefiting from the knowledge of the FM.
(2) \textit{Efficient Knowledge Aggregation}: the server integrates classifiers from multiple clients, facilitating knowledge transfer without access to private data.
In addition, we introduce novel regularization strategies that preserve activations of classes not present on the selected client and reduce common activations between similar classes.
Since the feature spaces of student and teacher models were aligned during pretraining, the aggregated classifier can be seamlessly plugged into the FM.
This innovative approach enables adaptation of a large-scale model to specific domains without direct access to training data, aligning with principles of modular deep learning by promoting composability, reusability, and personalization for privacy-preserving AI~\cite{pfeiffer2023modular}.
Our contributions are:
\begin{itemize}
    \item We introduce a novel paradigm, Cross-Architecture Federated Knowledge Transfer (CA-FKT), for FL across heterogeneous model architectures.
    \item We propose FedPromo, a scalable framework that leverages lightweight proxy models for training a large-scale model in a federated setting.
    \item We integrate cross-architectural knowledge distillation with federated optimization, addressing key challenges in personalization, generalization, and efficiency.
    \item We demonstrate the effectiveness of FedPromo across 5 image classification benchmarks, achieving state-of-the-art performance in privacy-preserving model adaptation.
\end{itemize}

\section{Related Works} \label{sec:related}
\noindent \textbf{Federated Learning (FL):}
Federated Learning (FL) enables decentralized learning without sharing private data, which is crucial for many vision tasks~\cite{shenaj2023federated}.
The first FL approach, \ie, FedAvg~\cite{mcmahan2017communication}, is effective in simple scenarios, but struggles with non-IID distributions.
FedProx~\cite{li2020federated} adds a proximal term to the local objectives to limit the impact of local updates, reducing clients' drift.
SCAFFOLD~\cite{karimireddy2020scaffold} estimates client drift by comparing server and client update directions and uses it for correction.
FedMargin~\cite{michieli2022federated} emphasizes the role of learned representations in model aggregation, while MOON~\cite{li2021model} improves local training in non-IID settings through model-based contrastive learning.

\noindent \textbf{Foundation Models (FMs):}
Foundation Models have advanced multiple domains, beginning with Natural Language Processing (\eg, BERT~\cite{devlin2019bert}) and extending to computer vision. CLIP~\cite{radford2021learningtransferablevisualmodels} marks a key development in Vision-Language Models (VLMs), employing contrastive learning to align image-text representations for strong zero-shot transfer. DINO~\cite{caron2021emerging} uses self-distillation to learn semantic object structures without supervision, improving representation learning for downstream vision tasks.
The rapid development of large-scale pretraining has resulted in FMs that can act as effective feature extractors for a variety of downstream tasks and domains.
In~\cite{ostapenko2022continual}, the authors explore the efficacy of pretrained vision models as a foundation for downstream continual learning.
FMs are also starting to gain interest in the context of FL. In \cite{liu2024fedfms}  the authors propose fine-tuning a pretrained SAM~\cite{sam} on each client and then introduce another approach of fine-tuning the parameters of the adapters and decoder of a medical SAM adapter.
In this work, we improve an FM in a scalable, modular, and privacy-preserving way, using FL on remote devices.

\noindent \textbf{Pretraining in FL:}
Traditional FL settings~\cite{mcmahan2017communication}, limit the role of the server as a sole aggregator. However, given its computational capabilities, it is advantageous to fully exploit it for improved efficiency and performance. For this reason, recent works have begun to explore pretraining steps at server-side before initiating local training on devices~\cite{shenaj2023federated}.
Pretraining allows us to better address data heterogeneity, allowing longer local training and reducing communication costs \cite{nguyen2023where, chen2023on}.
Beyond improving communication efficiency, pretraining enables more realistic FL scenarios by addressing tasks where client-side annotations are impractical. For example, pretraining on synthetically generated  supervised data facilitates unsupervised client-side learning for challenging vision tasks such as semantic segmentation \cite{shenaj2023learning}.

\noindent \textbf{Knowledge Distillation (KD):}
Originally proposed to transfer knowledge from large to small models~\cite{hinton2015distilling}, KD is widely used in FL to address model and data heterogeneity~\cite{mora2024knowledge}. Model-agnostic KD~\cite{jeong2018communication, li2019fedmd, hongyan2019cronus, wu2024exploring} improves FL without architectural limits, while data-agnostic KD~\cite{pmlr-v139-zhu21b, sattler2021fedaux, zhang2022fedzkt, yao2023fedgkd} boosts robustness under non-IID data. In contrast, we integrate KD into pretraining so lightweight models align with the server model before FL begins, reducing network load by sharing only a small decoder attached to both local and server models.

\section{Problem Setting} \label{sec:formulation}
In this section, we introduce the Cross-Architecture Federated Knowledge Transfer (\sname) task:
as in conventional FL settings, we assume a central server $s$ and a set of clients $\mathcal{K} = \{k\}_{k=1}^{K}$.
We assume that the client set $\mathcal{K}$ is divided into $G$ groups $\mathcal{K}^i, i=1,\dots,G$ such that $\mathcal{K} = \bigcup_{i} \mathcal{K}^i$ and $\mathcal{K}^i \cap \mathcal{K}^j = \varnothing, \forall i \neq j$.
Each group $\mathcal{K}^i$ learns a model $M^i$ from a private dataset $\mathcal{D}^{i}$.
For ease of notation, we focus on a single domain and omit the index $i$ in the following.

The FL process follows the typical communication scheme: clients perform local training, send their updated models to the server, which aggregates them, and redistributes the updated model. This cycle repeats for $R$ rounds.

In the conventional FL setup, the server and the clients train the same model architecture $M$, requiring similar computational resources for training.
Instead, we consider an encoder-decoder architecture for vision tasks and relax this assumption by considering the scenario where the clients and the server share only a simple decoder module $D$.
Formally, the server, with greater computational resources, maintains a large-scale model $M^s \!= \!D\! \circ \!O$ where $O$ (Oracle) is a large-scale encoder.
Clients, constrained by lower resources, use compact models $M^k = D^k \circ E$ where $E$ is a lightweight encoder.
The decoders are initialized as $D$, \ie, at the beginning of the federated optimization, $D^k=D, \forall k$, and then their optimization becomes the objective of the federated training.
All models share the same decoder structure $D$, ensuring compatibility, while the encoders are different (\ie,  $E\neq O$).

Given an input image $\mathbf{X} \in \mathcal{X} \subset \mathbb{R}^{H\times W\times 3}$, the encoders extract feature representations $\mathbf{O} = O(\mathbf{X})$ and $\mathbf{F} = E(\mathbf{X})$, where $\mathbf{O}, \mathbf{F} \in \mathbb{R}^F$.
Importantly, to enable seamless integration of client-side models with the shared decoder, the features extracted on the client-side must approximate those of the server, that is, $\mathbf{F} \simeq \mathbf{O}$ for any given input $\mathbf{X}$.
To achieve this, we exploit a translator block at client-side and cross-architectural distillation during pretraining on samples from an auxiliary dataset $\mathcal{D}^s$ prior to federated optimization (Sec. \ref{subsubsec:pretrain}).
The objective of the \sname~task is to enable the server to classify data from the same domain as the one privately attained by the clients $\mathcal{D} = \bigcup_{k=1}^{K} \mathcal{D}_k$ without direct access to client data.
To ensure privacy and efficiency, clients train only the decoder $D$, which is task-specific and operates on high-level representations~\cite{sidahmed2021efficientprivatefederatedlearning,Chen_Zhang_Krompass_Gu_Tresp_2024,9546506}.
In the multi-domain scenario, after training, the domain-specific decoders $D^i$ can be concatenated together on the server and  attached to the Oracle encoder, enabling recognition in all domains with minimal performance loss (Sec.~\ref{sec:results}).

\begin{figure*}[t]
    \centering
    \includegraphics[width=.83\textwidth]{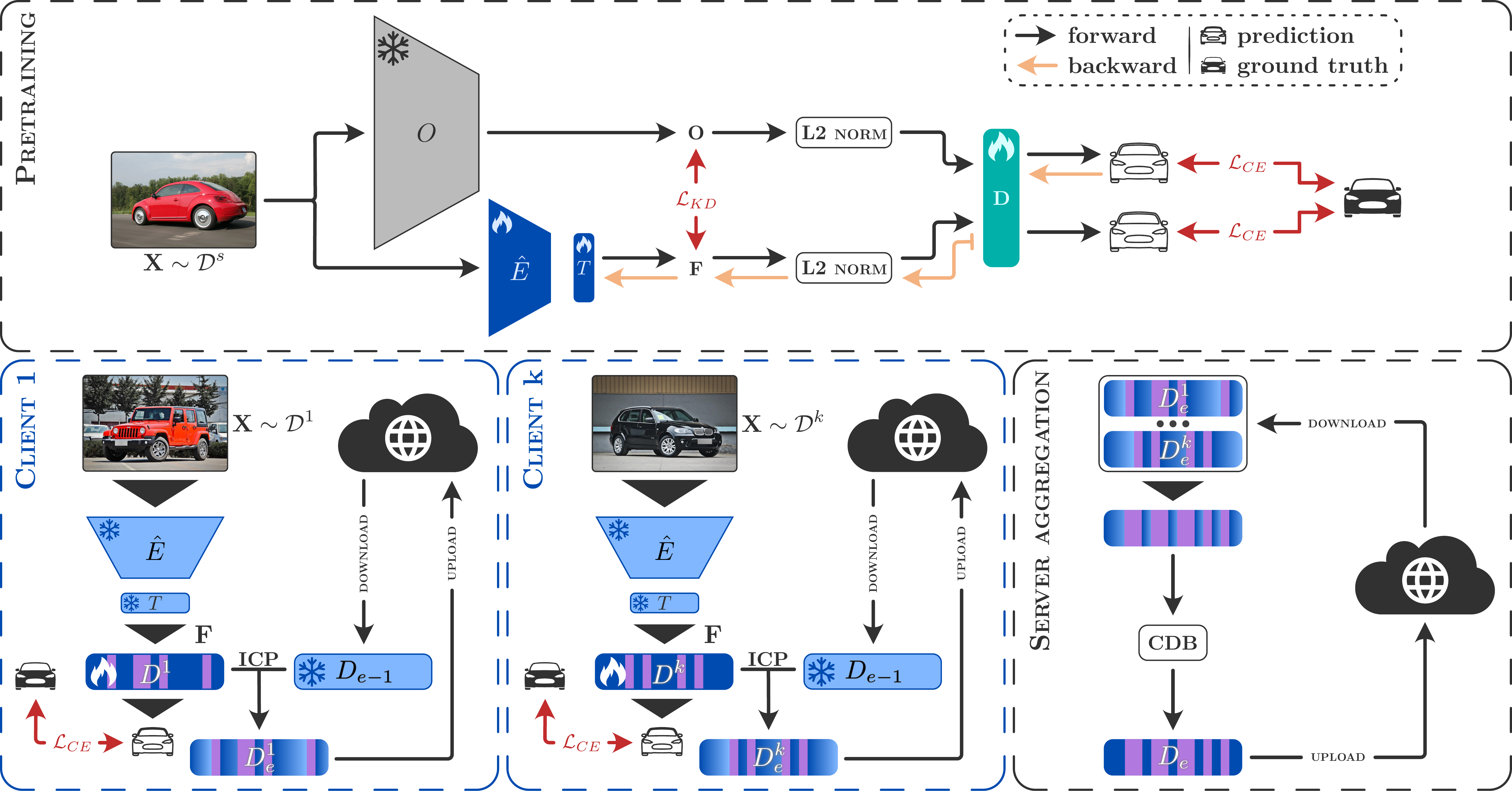}
    \caption{FedPromo employs a two-stage architecture. The top row illustrates the cross-architectural distillation pretraining, where we transfer features useful for classification by training the decoder $D$ only on Oracle features, and keeping it frozen when the student feature vector $\mathbf{F}$ is in input. We also add an extra knowledge distillation loss acting directly on the features. The bottom row summarizes the federated setup, depicting decentralized training on two sample clients and subsequent model aggregation on the server.}   \label{fig:arch}
\end{figure*}

\section{Proposed Method} \label{sec:method}

\subsection{The FedPromo Framework} \label{subsec:arch}
The framework is built on an encoder-decoder classification network. Its key innovation is supporting different encoders on the server and client sides while maintaining a common feature space, enabling a shared decoder across both environments. This approach bridges the computational gap between server and client. The server capitalizes on higher computing capacity with a computationally intensive, yet highly capable, encoder network. In contrast, at client-side, with limited computational and energy resources,  a significantly smaller and lighter network is used as a proxy for the server counterpart.
In our experiments, we used DINOv2~\cite{oquab2023dinov2} based on ViT-L/14-Reg~\cite{darcet2023vitneedreg} at server side due to its ability to extract rich features across domains, while clients use a MobileNetV3-small~\cite{howard2019searching}, though any foundation and lightweight models can be used.

To allow cross-architectural distillation and enable decentralized training of a shared classifier, a translator block $T\!: \! \mathbb{R}^{F'} \! \! \mapsto \! \mathbb{R}^F$ that maps the feature space of the client encoder to that of the server is needed.
In this setting $E = T \circ \hat{E}$, and $\hat{E}: \mathcal{X} \rightarrow \mathbb{R}^{F'}$.
In our implementation, we use the concatenation of the DINOv2 class token as the feature space, with the global average pooling of patch tokens~\cite{barbato2024cross}, and we implement $T$ as a linear transformation for efficiency.
The feature dimensions are $F' = 576$ (MobileNet) and $F = 2048$ (DINOv2). However, any translator module suitable for the selected networks can be used.
The translator is pretrained and remains frozen during federated optimization, ensuring consistent feature representations across clients and the server.

To minimize client-side computation, we use a single linear layer as the classifier, defined as \mbox{$D: \mathbb{R}^F \rightarrow [0,1]^{C}$}, mapping the shared feature space to a discrete probability distribution over the $C$ classes of the considered domain, where $C = |\mathcal{C}|$.
Additionally, to regularize the gradients during federated training, we introduce an L2-normalization layer between the encoders and the decoder, leading to improved performance (see the \textit{Supp. Mat.}).
For reference, in our setting, the client-side model (encoder + translator + classifier) has 100× fewer parameters than the server model. Fig.~\ref{fig:arch} visually illustrates our architecture and the interactions between server and client components.

\subsection{Training Procedure}
FedPromo requires aligned feature spaces between server and client encoders to enable a shared decoder. We achieve this through cross-architectural distillation during pretraining, followed by federated training on private client data.

\subsubsection{Cross-Architecture Pretraining} \label{subsubsec:pretrain}
Our pretraining strategy couples the primary classification task with cross-architectural distillation~\cite{barbato2024cross}. The global optimization objective is $\mathcal{L}= \mathcal{L}_{KD} + \lambda \mathcal{L}_{CE}$, where $\mathcal{L}_{CE}$ is the categorical cross-entropy loss, and $\mathcal{L}_{KD}$ is the knowledge distillation loss for feature reconstruction.
We empirically set $\lambda \!=\! 0.5$.
The $\mathcal{L}_{KD}$ term enforces feature alignment between student (client) and teacher (server) encoders by aligning both the norm and the orientation of the feature vectors in the high-dimensional embedding space.
Formally, for a given training sample $\mathbf{X}\in \mathcal{D}^s$, the knowledge distillation loss can be expressed as:
\begin{equation}
\mathcal{L}_{KD} = \ell_1 \left(\mathbf{F},\mathbf{O}\right) + \ell_2 \left(\mathbf{F},\mathbf{O}\right) + \ell_{cos} \left(\mathbf{F},\mathbf{O}\right) \;,
\end{equation}
where $\ell_1$, $\ell_2$ and $\ell_{cos}$ represent the corresponding distances.

For best performance, the dataset $\mathcal{D}^s$, where the cross-entropy loss is optimized, should be in-domain to the subsequent federated training data.
To reinforce distillation, we train the classifier only on frozen teacher features and apply it directly to the student. This forces the student to replicate the most relevant features for classification according to the server model, improving accuracy (see Secs.~\ref{sec:results} and \ref{sec:ablation}).

\subsubsection{Federated Learning}
\label{subsubsec:federated}
The complexity of the CA-FKT scenario requires several novel contributions to the federated optimization process. We focus on regularizing the client-side training to maintain a high degree of alignment between client and server behaviors. This is crucial when considering a realistic scenario where the federated partition is highly non-IID, resulting in minimal class overlap among clients.

\noindent We introduce two key innovations to address these issues:
\begin{enumerate}
\item A client-side regularization objective, Inactive Classes Preservation (ICP), which, drawing inspiration from continual learning~\cite{wang23survey,barbato2024continual}, preserves the knowledge of classes not present in the client data, that cross-entropy tends to diminish.
\item A weight filtering strategy,  called Class De-Biasing (CDB), is implemented at both  client and server-side: it removes the average output activation for each class from the classifier, thereby reducing overall confusion.
\end{enumerate}
See the \textit{Supp. Mat.} for a pseudocode of the implementation. 

\noindent\textbf{Inactive Classes Preservation (ICP):}
Our federated scenario also allows the learning of fine-grained datasets partitioned non-IID across clients. 
This setup leads to significant performance degradation when using cross-entropy loss as the sole optimization objective: since each client $k$ contains a limited number of classes, it can easily achieve a lower entropy score (and therefore loss) by minimizing the predicted probability of classes absent from its set.
Over time, this effect leads to overconfident predictions of frequent classes, even after aggregation.

We address this issue by implementing selective weight optimization for the clients. Essentially, each client keeps track of the labels encountered at each local epoch and enforces smaller updates on the classifier weights corresponding to the missing classes.
Let us define $D_e^k$ the decoder of client $k$ at the end of current local epoch $e$ (mapping features of size $F$ to $C$ classes), its weights as $\boldsymbol{\omega}_e \in \mathbb{R}^{C \times F}$, and $\mathcal{A} \subset \mathcal{C}$ the subset of active classes (\ie, those seen by the client). We can express the regularization strategy as:
\begin{equation}
    \boldsymbol{\omega}_{e}^{\text{ICP}}[c] = \begin{cases}
    \eta \boldsymbol{\omega}_{e}[c] + (1-\eta) \boldsymbol{\omega}_{e-1}[c] & \forall c \in \mathcal{C} \setminus \mathcal{A} \\
    \boldsymbol{\omega}_{e}[c] & \forall c \in  \mathcal{A}
    \end{cases} \;,
\end{equation}
where $\eta$ has been empirically tuned to $0.05$ (see \textit{Supp. Mat.} for details), and $\mathcal{C} \setminus \mathcal{A}$ is the set of inactive classes.
Note that, at the beginning, we initialize the weights from the pretraining output model received from the server.

\noindent\textbf{Class De-Biasing (CDB):}
Our second contribution focuses on reducing common activations between classes in the classifier weights, which leads to lower confidence and error-prone predictions. Given the fine-grained nature of our setting, where classes often share many similarities (\eg, all vehicles considered in CompCars are, in fact, cars), it is crucial to ensure that the classifier is not affected by the \textit{bias} component common to all classes of the domain. This means that it should focus on distinguishing between the fine-grained classes, rather than categorizing them as their superclass. For example, in CompCars, it should focus on recognizing how each car differs from others, rather than on recognizing them all as cars.
We perform %
de-biasing both locally at the clients and globally at the server.
This approach %
reduces cross-talk between %
classes, %
avoiding common feature activation from %
visually similar classes (see  Fig.~\ref{fig:CDB_sim} in the ablation).
Formally, before the aggregation step, each active client $k \in \mathcal{K}_A$ ($K_A = |\mathcal{K}_A|$) updates %
its classifier weights $\boldsymbol{\omega}^k_r \!= \! \boldsymbol{\omega}^{\text{ICP}}_{e_{\text{last}}}$ ($e_{\text{last}}$ is the last local epoch) as: %
\begin{equation}
    \boldsymbol{\omega}_{r+1}^k[c] = \boldsymbol{\omega}_r^k[c] - \frac{1}{C} \sum_{c=1}^C \boldsymbol{\omega}_r^k[c] \;\; .
\end{equation}
This newly debiased weight is then sent to the server for aggregation, which performs the same operation on the weights $\boldsymbol{\omega}^s_{r+1}$  obtained after aggregation at the server-side.

\section{Experimental Results} \label{sec:results}
\noindent \textbf{Implementation Details:}
We implemented our approach using the \textit{Flower} framework~\cite{beutel2020flower} in PyTorch.
For pretraining, we train our architectures on the server for 60 epochs using a batch size of 64. We adopt a cosine-annealing learning rate scheduler (decaying to $lr=0$) with a peak learning rate of ${lr}_{\text{max}} = 0.005$. Optimization is performed using Adam ($\beta_1 = 0.9$, $\beta_2 = 0.999$) with no weight decay. We follow the TorchVision benchmark~\cite{torchvision2016} for MobileNet data augmentation.
For federated learning, we keep the optimizer, scheduler, batch size, and peak learning rate the same as in pretraining. The optimization runs for $R = 500$ rounds, with $e_{\text{max}} = 10$ local epochs per client (without augmentation). We consider a federated setup with $K = 100$ clients, where $K_A = 10$ clients participate per-round. %
Pretraining requires 40 minutes on an NVIDIA L40s for StanfordCars and $\sim$2 days for ImageNet. Federated finetuning on CompCars requires $\sim$9 hours for 500 rounds; timing on other datasets scales with size (see \textit{Supp. Mat.} for details).

\noindent \textbf{Datasets:}
We evaluate FedPromo on five domain pairs exhibiting long-tailed class distributions (average of 272 target classes, all belonging to similar objects) distributed on the clients according to a Dirichlet distribution with concentration parameter $\alpha = 1$~\cite{hsu2019measuringeffectsnonidenticaldata}. We include a representation of the employed splits in the \textit{Supp. Mat.}.
In each scenario, a public source dataset is used for pretraining, while a private target dataset is distributed across clients in a federated setting.
We consider two pairs of datasets containing images of animals (CUB200 \cite{wah2011caltech} $\rightarrow$ NABirds \cite{Horn_2015_CVPR}, ImageNetPets $\rightarrow$ OxfordPets \cite{parkhi2012cats}), two pairs containing images of vehicles (StanfordCars~\cite{krause20133d}~$\rightarrow$~CompCars~\cite{yang2015large}, FGVCAircraft~\cite{maji13fine-grained}~$\rightarrow$~MilitaryAircraft~\cite{militaryAricraft}) and one pair containing images of food (Food101~\cite{bossard2014food}~$\rightarrow$~UECFOOD256~\cite{kawano14c}). 
The only custom dataset is \textit{ImageNetPets}, that  we extracted from the ImageNet-1k dataset~\cite{imagenet_cvpr09} using the WordNet hierarchy \cite{christiane2005wordnet} (we will release the split with the codebase, see the \textit{Suppl.~Mat } for details). %
The source domain is considered public, allowing the server to pretrain a proxy model. This proxy model is then used by clients to learn knowledge from their private target domains in a federated manner. Results without pretraining dataset are also provided. See the \textit{Supp.\ Mat.\ }for more details on the datasets and clients' splits. %

\noindent \textbf{Experimental Comparison:}
We compare FedPromo against four baselines using Top-1 and Top-5 classification accuracy. We evaluate for reference both the proxy models (MobileNet) and the Oracle model (DINOv2), where the federated decoder is attached to the server encoder.
Firstly, we consider the common FedAvg~\cite{mcmahan2017communication} strategy and its extension with an exponential moving average, FedAvg + EMA (with rate $0.5$).
We then evaluate two advanced federated strategies, FedProx~\cite{li2020federated} and MOON~\cite{li2021model}.
The former adds a regularization term to clients during training, named \textit{proximal term}, slowing the evolution of the training but achieving greater stability. We consider the best-case scenario where no straggler clients are present in FedProx \cite{li2020federated}.
The latter extends FedProx with knowledge distillation on network outputs and contrastive learning for feature alignment.
Finally, for an upper-bound reference, we report the \textit{Centralized} setup, where classifiers are trained using standard non-federated optimization.
Note that our centralized setup is meant to reflect the federated scenario, so only the decoder is trained, and the features in input are those of the proxy model.
Centralized training runs for 50 epochs, matching the total iteration count of the FL setup.

\begin{table}[t]
    \renewcommand{\arraystretch}{.9}
    \centering
    \small
    \begin{tabular}{c|c|c:c|c:c}
       \multirow{2}{*}{$\mathcal{D}$} & \multirow{2}{*}{Method} & \multicolumn{2}{c|}{MobileNet} & \multicolumn{2}{c}{DINOv2} \\
       & & Top-1 &  Top-5 &  Top-1 &  Top-5 \\
       \hline
       \multirow{7}{*}{\rotatebox{90}{CompCars}} & FedAvg & \underline{26.4} & \underline{43.4} & 14.0 & 27.4 \\
       & FedAvg + EMA & 18.5 & 35.9 & \underline{21.5} & \underline{42.6} \\
       & FedProx${}^\dagger$ & 9.5 & 24.2 & 13.7 & 31.2 \\
       & MOON${}^\dagger$ & \underline{26.4} & 43.3 & 13.8 & 27.4 \\
       & FedPromo (ours) & \textbf{37.2} & \textbf{63.5} & \textbf{35.8} & \textbf{69.4} \\
       \cdashline{2-6}
       & Centralized & 48.3 & 75.9 & 41.1 & 73.2 \\
       \hline
       \multirow{7}{*}{\rotatebox{90}{UECFOOD256}} & FedAvg & \underline{48.6} & \underline{74.4} & 29.6 & 50.4 \\
       & FedAvg + EMA & 41.4 & 73.0 & \underline{49.1} & \underline{83.1} \\
       & FedProx${}^\dagger$ & 5.6 & 30.1 & 8.2 & 42.4 \\
       & MOON${}^\dagger$ & 48.4 & 74.3 & 29.7 & 50.5 \\
       & FedPromo (ours) & \textbf{52.1} & \textbf{80.0} & \textbf{58.7} & \textbf{88.7} \\
       \cdashline{2-6}
       & Centralized & 55.3 & 83.0 & 60.1 & 88.3 \\
       \hline
       \multirow{7}{*}{\rotatebox{90}{NABirds}} & FedAvg & 23.6 & 41.2 & 6.1 & 12.9 \\
       & FedAvg + EMA & 20.7 & 41.1 & \underline{20.7} & \underline{43.6} \\
       & FedProx${}^\dagger$ & 12.8 & 31.4 & 16.0 & 39.9 \\
       & MOON${}^\dagger$ & \underline{23.8} & \underline{41.4} & 6.1 & 12.9 \\
       & FedPromo (ours) & \textbf{32.2} & \textbf{59.0} & \textbf{30.3} & \textbf{65.0} \\
       \cdashline{2-6}
       & Centralized & 35.3 & 62.8 & 32.3 & 65.3 \\
       \hline
       \multirow{7}{*}{\rotatebox{90}{MilitaryAircraft}} & FedAvg & \underline{10.7} & \textbf{29.3} & 9.2 & 27.7 \\
       & FedAvg + EMA & 8.1 & 26.7 & \textbf{12.4} & \textbf{39.0} \\
       & FedProx${}^\dagger$ & 6.9 & 23.5 & 10.4 & \underline{33.8} \\
       & MOON${}^\dagger$ & \textbf{10.8} & \underline{29.1} & 9.6 & 27.8 \\
       & FedPromo (ours) & 9.9 & \textbf{29.3} & \underline{11.9} & 32.1 \\ %
       \cdashline{2-6}
       & Centralized & 11.3 & 31.5 & 13.4 & 35.4 \\
       \hline
       \multirow{7}{*}{\rotatebox{90}{OxfordPets}} & FedAvg & \underline{56.9} & \underline{89.1} & 47.9 & 79.2 \\
       & FedAvg + EMA & 52.4 & 87.7 & 67.6 & \underline{96.4} \\
       & FedProx${}^\dagger$ & 54.1 & \underline{89.1} & \underline{68.0} & 95.0 \\
       & MOON${}^\dagger$ & 56.5 & 88.8 & 47.4 & 78.8 \\
       & FedPromo (ours) & \textbf{59.2} & \textbf{91.6} & \textbf{72.7} & \textbf{98.0} \\
       \cdashline{2-6}
       & Centralized & 61.6 & 92.2 & 71.2 & 94.4 \\
    \end{tabular}%
    \caption{Federated Top-1 and Top-5 accuracy ($\uparrow$) when optimization starts from an in-domain dataset.
    ${}^\dagger$strategy partially re-implemented to adapt to our specific settings. Evaluation is done on the server-side after federated aggregation. \textbf{Best} in bold, \underline{second-best} underlined. %
    }
    \label{tab:results}
\end{table}

\begin{table}[hpt]
    \renewcommand{\arraystretch}{.9}
    \newcommand{\hpad}{\hspace*{0em}}
    \centering
    \small
    \begin{tabular}{c|c|c:c|c:c}
        \multirow{2}{*}{$\mathcal{D}$} & \multirow{2}{*}{Method} & \multicolumn{2}{c|}{MobileNet} & \multicolumn{2}{c}{DINOv2} \\
         & & \hpad Top-1\hpad & \hpad Top-5\hpad & \hpad Top-1\hpad & \hpad Top-5\hpad \\
        \hline
        \multirow{7}{*}{\rotatebox{90}{CompCars}} & FedAvg & \underline{16.2} & 31.4 & 1.2 & 3.9 \\
        & FedAvg + EMA & 6.5 & 17.5 & \underline{3.0} & \underline{9.7} \\
        & FedProx${}^\dagger$ & 2.1 & 7.6 & 2.4 & 8.5 \\
        & MOON${}^\dagger$ & \underline{16.2} & \underline{31.5} & 1.3 & 3.9 \\
        & FedPromo (ours) & \textbf{18.3} & \textbf{37.1} & \textbf{3.3} & \textbf{10.8} \\ %
        \cdashline{2-6}
        & Centralized & 29.7 & 53.1 & 3.4 & 9.5 \\
        \hline
        \multirow{6}{*}{\rotatebox{90}{UECFOOD256}} & FedAvg & 43.2 & \underline{68.9} & 2.1 & 6.9 \\
        & FedAvg + EMA & 36.2 & 64.5 & \underline{23.5} & \underline{48.4} \\
        & FedProx${}^\dagger$ & 3.0 & 13.0 &  4.0 & 13.2 \\
        & MOON${}^\dagger$ & \underline{43.8} & 68.4 & 2.0 & 7.1 \\
        & FedPromo (ours) & \textbf{48.2} & \textbf{75.8} & \textbf{24.0} & \textbf{48.5} \\ %
        \cdashline{2-6}
        & Centralized & 51.8 & 78.4 & 27.0 & 56.5 \\
        \hline
        \multirow{7}{*}{\rotatebox{90}{NABirds}} & FedAvg & \underline{28.0} & \underline{46.7} & 1.5 & 5.1 \\
        & FedAvg + EMA & 21.2 & 41.8 & \underline{11.3} & \underline{29.1} \\
        & FedProx${}^\dagger$ & 10.3 & 25.7 & 10.5 & 26.6 \\
        & MOON${}^\dagger$ & \underline{28.0} & 46.5 & 1.6 & 5.0 \\
        & FedPromo (ours) & \textbf{37.5} & \textbf{65.1} & \textbf{18.5} & \textbf{48.6} \\ %
        \cdashline{2-6}
        & Centralized & 42.7 & 70.5 & 17.3 & 47.6 \\
        \hline
        \multirow{6}{*}{\rotatebox{90}{MilitaryAircraft}} & FedAvg & 16.4 & 38.1 & 5.1 & 16.2 \\
        & FedAvg + EMA & 10.6 & 28.9 & \underline{8.0} & \textbf{25.8} \\
        & FedProx${}^\dagger$ & 8.8 & 24.2 & \textbf{8.3} & \underline{24.8} \\
        & MOON${}^\dagger$ & \underline{16.5} & \underline{38.2} & 5.3 & 16.2 \\
        & FedPromo (ours) & \textbf{16.9} & \textbf{40.0} & 7.1 & 24.1 \\ %
        \cdashline{2-6}
        & Centralized & 18.9 & 43.4 & 6.7 & 25.9 \\
        \hline
        \multirow{7}{*}{\rotatebox{90}{OxfordPets}} & FedAvg & \textbf{82.4} & 97.2 & 84.0 & 99.5 \\
        & FedAvg + EMA & 77.1 & \underline{97.4} & 85.2 & \underline{99.8} \\
        & FedProx${}^\dagger$ & 78.2 & \underline{97.4} & \underline{85.6} & \textbf{99.9} \\
        & MOON${}^\dagger$ & \textbf{82.4} & \underline{97.4} & 84.7 & 99.5 \\
        & FedPromo (ours) & \underline{82.2} & \textbf{98.2} & \textbf{89.5} & \underline{99.8} \\ 
        \cdashline{2-6}
        & Centralized & 83.7 & 98.2 & 87.8 & 99.8 \\
    \end{tabular}
    \caption{Experimental results when federated optimization starts from an out-of-domain dataset. Federated Top-1 and Top-5 accuracy ($\uparrow$) starting from ImageNet-1k (1000 classes) pretraining. ${}^\dagger$strategy partially re-implemented to adapt to our specific settings. \textbf{Best} in bold, \underline{second-best} underlined.
    }
    \label{tab:imagenet}
\end{table}

\noindent \textbf{ Results with In-Domain Pretraining:}
The results in Tab. \ref{tab:results} show the strong and consistent performance of FedPromo across various scenarios. Our approach outperforms existing methods in most cases, achieving the highest Top-1 accuracy in 8 out of 10 scenarios and the best Top-5 accuracy in 9 of them. %
In the remaining cases, FedPromo typically ranks second, demonstrating its robustness and stability.

In particular, improvements over existing methods are often substantial. For example, in the StanfordCars$\rightarrow$CompCars setup, FedPromo achieves a Top-1 accuracy of $37.2\%$ for the proxy models, outperforming the next best method by $10.8$ percentage points.
Similarly, in the Food101$\rightarrow$UECFOOD256 scenario, our approach reaches $58.7\%$ Top-1 accuracy when applied to DINOv2 (\textit{server} model), outperforming the second-best method (FedAvg + EMA) by $9.6$ percentage points.
The gain is also evident in the CUB200$\rightarrow$NABirds setting, where FedPromo achieves a server Top-1 accuracy of $30.3\%$, compared to $16.0\%$ of FedProx and $20.7\%$ of FedAvg + EMA.
These results highlight FedPromo’s advantage over competing techniques, particularly when leveraging the server-side DINOv2 model.
A similar improvement is also notable in the $\text{ImageNetPets}\rightarrow\text{OxfordPets}$ setting, where our approach is consistently the best performer, achieving an average accuracy improvement of $2.8\%$.

A key strength of FedPromo in the CA-FKT task is its consistency across different architectures. Our approach delivers top-tier performance for both MobileNet (proxy) and DINOv2 (server), demonstrating: (i) versatility across different network designs, and (ii) robust feature alignment through effective pretraining.
Even in the most challenging case (FGVCAircraft$\rightarrow$MilitaryAircraft), FedPromo remains on par with the best competitors, ensuring reliable performance across diverse domains.
Compared to the upper-bound (centralized training), FedPromo shows significant promise.
On proxy models, it achieves an average gap of just $4.8\%$ (Top-1) and $5.1\%$ (Top-5).
On server models, this difference shrinks to $0.3\%$ (Top-1) and $4.9\%$ (Top-5), approaching centralized performance.

\noindent \textbf{Results with Out-Of-Domain Pretraining:}
To further assess the generalization of FedPromo, we evaluate its performance using ImageNet-1k~\cite{imagenet_cvpr09} for pretraining, without any in-domain pretraining dataset (see Tab. \ref{tab:imagenet}).
FedPromo remains highly effective even without in-domain pretraining, reinforcing its adaptability.
Our approach achieves performance very close to centralized training, both on the proxy model and the server model.
It outperforms competitors on both Top-1 and Top-5 in 8 out of 10 settings, getting very close to the best in the remaining ones.
In particular, performance gains are substantial on the \textit{NABirds}, \textit{UECFOOD}, and \textit{CompCars} datasets, while the MilitaryAircraft is challenging for all approaches, and performance on OxfordPets tends to saturate, making it difficult to see the difference since ImageNet contains many of the target classes.

\begin{table}[t]
    \renewcommand{\arraystretch}{.9}
    \centering
    \begin{tabular}{c|c:c|c:c}
        \multirow{2}{*}{Dataset} & \multicolumn{2}{c|}{Specific} & \multicolumn{2}{c}{Agnostic} \\
        & Top-1 & Top-5 & Top-1 & Top-5 \\
        \hline
        CompCars & 35.8 & 69.4 & 35.8 & 69.1 \\
        UECFOOD256 & 58.7 & 88.7 & 58.7 & 88.4 \\
        NABirds & 30.3 & 65.0 & 30.2 & 64.8 \\
        Military Aircraft & 11.9 & 32.1 & 4.6 & 8.9 \\
        OxfordPets & 72.7 & 98.0 & 70.9 & 92.8 \\
        \hline
        Average & 41.9 & 70.6 & 35.4 & 65.4 \\
    \end{tabular}
    \caption{Experiments on multi-domain scenario, server accuracy. \textit{Specific}: classifiers are kept separated (requires samples' domain knowledge). %
    \textit{Agnostic:} %
    classifiers are concatenated, unknown samples %
    (no domain information assumed).}
    \label{tab:multi_domain}
\end{table}

\begin{table}[t]
    \renewcommand{\arraystretch}{.9}
    \centering
    \begin{tabular}{cc|c:c|c:c|c}
        \multirow{2}{*}{CDB} & \multirow{2}{*}{ICP} & \multicolumn{2}{c|}{MobileNet} & \multicolumn{2}{c}{DINOv2} & \multirow{2}{*}{Avg.}\\
        & & Top-1 & Top-5 & Top-1 & Top-5 \\
        \hline
        \xmark & \xmark & 26.4 & 43.4 & 14.0 & 27.4 & 27.8 \\
        \xmark & \cmark & 36.8 & 63.3 & 35.6 & 69.2 & 51.2 \\
        \cmark & \cmark & 37.2 & 63.5 & 35.8 & 69.4 & 51.5 \\
    \end{tabular}%
    \caption{Method components ablation. Federated Top-1 and Top-5 accuracy ($\uparrow$) on the CompCars (431 classes) dataset starting from StanfordCars (196 classes) pretraining.}
    \label{tab:ablation_comp}
\end{table}

\noindent \textbf{Multi-Domain Results:}
Tab.~\ref{tab:multi_domain} reports the accuracy achieved by the DINOv2 model when classifying data from all the domains together after distributed optimization using FedPromo. We compare the domain-\textit{specific} accuracy (same configuration as Tab.~\ref{tab:results}) with that of the domain-\textit{agnostic} approach, where we concatenate all the classifiers of the single domains into a single larger one.
In the latter case, our method achieves competitive results, closely matching the domain-specific performance.
The average loss of top-1 accuracy across domains is just $3.2\%$, while for top-5  the accuracy loss is $3.6\%$. Moreover, note that most of the performance degradation arises from the MilitaryAircraft dataset, where the low starting accuracy led to confusion with classes in the other domains when concatenating the classifiers.
More details are in the \textit{Supp.~Mat.}

\section{Ablation} \label{sec:ablation}
In this section, we present an extensive ablation study, analyzing the contribution of each component, and discussing some limitations of our approach. We report ablation experiments in the StanfordCars$\rightarrow$CompCars scenario. This domain was chosen for its mid-range accuracy among the datasets in Tab.\ref{tab:results}, making it a balanced test case.

\noindent \textbf{Method Components' Ablation:}
Tab. \ref{tab:ablation_comp} presents the results of our component ablation study, in which we investigate the effect of the CDB and ICP modules in our federated approach.
The baseline performance without both of them  is relatively low for both proxy and server models.
Adding the ICP module results in a significant accuracy boost.
This is further strengthened by the inclusion of CDB, which achieves the best performance in all cases, in spite of the minimal computational cost.
These results confirm that both components effectively contribute to our approach.

\begin{figure}[t]
    \centering
    \includegraphics[width=0.95\linewidth]{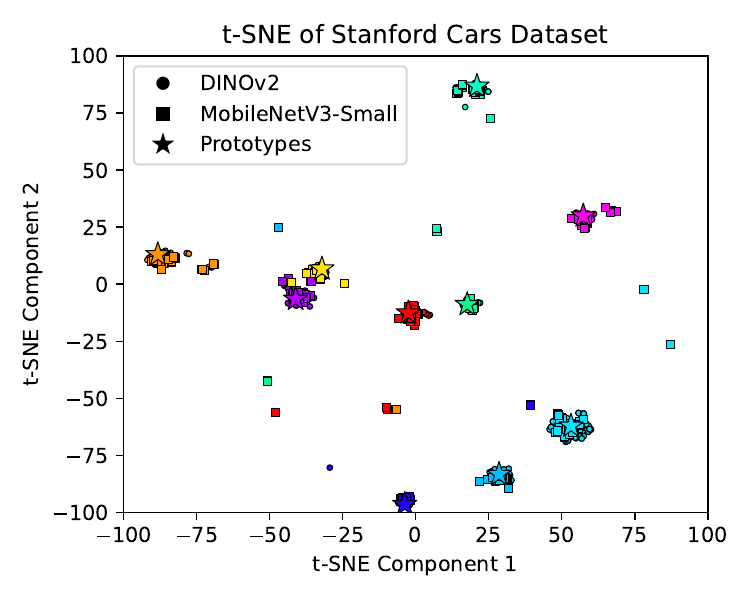}
    \caption{t-SNE plot showing the alignment of the features after pretraining.
    For clarity, it shows the embeddings of just 10 classes. The full plot with all classes is in the \textit{Supp. Mat.}
    }
    \label{fig:tsne}
\end{figure}

\noindent \textbf{Feature Alignment:}
Fig. \ref{fig:tsne} shows a t-SNE plot of the feature alignment achieved by our pretraining strategy.
This plot clearly illustrates how the features of DINO ($\circ$) and MobileNet ($\square$) are well-aligned class-by-class and clustered in the same regions.
Furthermore, the DINO class prototypes ($\medwhitestar$) are located close to their respective feature clusters, further confirming the alignment.
For better clarity, Fig. \ref{fig:tsne} presents only a subset of classes. A full visualization is available in the \textit{Supp.~Mat.}, where our findings are confirmed.
These results highlight the effectiveness of our pretraining strategy in aligning features across models, which is critical for federated knowledge transfer.

\noindent \textbf{$\mathbf{(\epsilon,\delta)}$ Differential Privacy:}
To study resilience against privacy attacks, we follow the \mbox{$(\epsilon,\delta)$-Differential Privacy} framework \cite{abadi2016deep,wei2020federated}.
Tab.~\ref{tab:dp_epsilon} reports the experiments on the CompCars dataset, where we fix $\delta\!=\!1e\!-\!5$ and vary $\epsilon$.
FedPromo performs well even at very low values of $\epsilon$, which correspond to a very high overall privacy budget. Indeed, even at $\epsilon\!=\!5$, the average accuracy is $46.7\%$, that is just $4.8\%$ less than the setting without DP. %

\noindent \textbf{Different Federated Training Configurations:}
To further analyze the stability of the training, we performed additional experiments on the CompCars dataset, varying the number of clients, the probability of transmission failure, and the Dirichlet parameter $\alpha$.
With fewer total clients ($K=50$) the average accuracy improved by $4.1\%$, while with more clients ($K=150$) it decreased by $9.4\%$; due to increased data heterogeneity and reduced class coverage on each client.
Similarly, decreasing the number of active clients per-round to $K_A=5$ leads to an improvement in accuracy of $0.7\%$, while increasing it to $K_A=20$ leads to a decrease of $3.3\%$.
To assess robustness, we simulated random uplink (client$\rightarrow$server) transmission dropout with probabilities $10\%$, $30\%$, and $50\%$, observing an average accuracy degradation of $\sim\!0.1\%$ in all cases.
Finally, varying the Dirichlet parameter we observe that, on average, every time $\alpha$ is decreased by $10\times$, one incurs a relative drop in accuracy by $\sim\!37.2\%$. A similar analysis on FedAvg highlights a higher drop of $\sim\!49.8\%$.

\noindent \textbf{Strength of ICP Regularization:}
Tab.~\ref{tab:eta} reports a study on the parameter $\eta$ that controls the update of inactive classes in the ICP block. Completely disabling updates on inactive classes ($\eta = 0$) results in subpar performance.
On the other side, using a large value ($\eta = 0.5$) causes an even stronger accuracy degradation. We experimentally found that setting $\eta=0.05$ provides the highest overall accuracy.

\begin{table}[t]
    \renewcommand{\arraystretch}{.9}
    \centering
    \begin{tabular}{c|c:c|c:c}
       \multirow{2}{*}{{\large$\epsilon$}} & \multicolumn{2}{c|}{Client} & \multicolumn{2}{c}{Server} \\
       & Acc-1 & Acc5 & Acc-1 & Acc-5 \\
       \hline
       \xmark & 37.2 & 63.5 & 35.8 & 69.4 \\
       \hline
       50 & 35.3 & 61.5 & 34.7 & 67.8 \\
       10 & 34.9 & 61.1 & 33.6 & 66.6 \\
       5 & 33.7 & 59.8 & 31.0 & 62.4 \\
       2.5 & 30.4 & 56.4 & 24.7 & 54.0 \\
       1 & 19.8 & 42.0 & 14.2 & 33.7 \\
    \end{tabular}
    \caption{Differential Privacy \cite{abadi2016deep} experiments varying $\epsilon$ with $\delta=10^{-5}$.
    More details on the hyperparameters used for differential privacy in the \textit{Supp.~Mat.}.}
    \label{tab:dp_epsilon}
\end{table}

\begin{figure}[t]
    \centering
    \includegraphics[width=\linewidth,trim=1.5cm 0.8cm 0.8cm 1cm, clip]{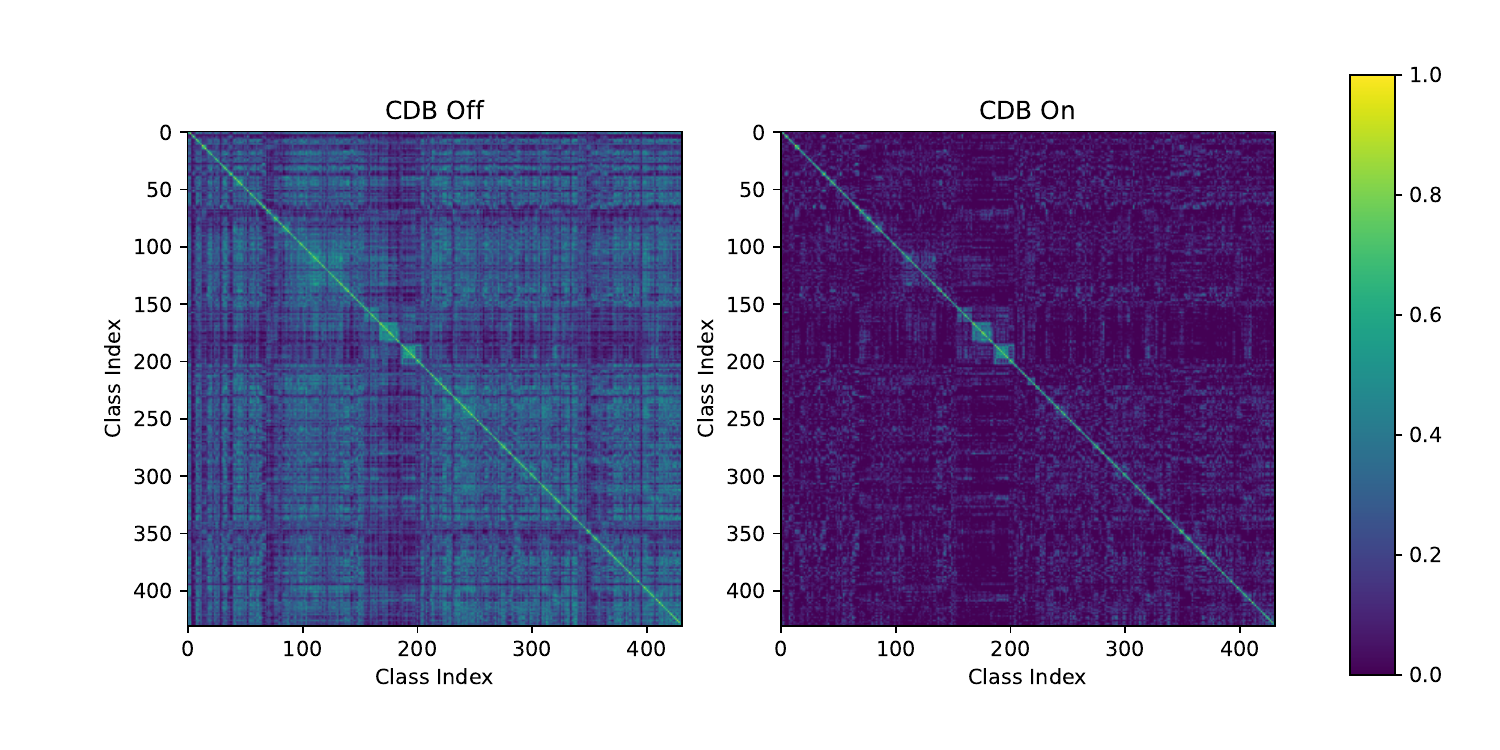}
    \caption{Self-similarity plots of the CompCars classifier weights trained with CDB off (left) and with CDB on (right). Our CDB block drastically reduces cross-talk between classes.}
    \label{fig:CDB_sim}
\end{figure}

\noindent \textbf{Effect of CDB:}
Fig.~\ref{fig:CDB_sim} illustrates the impact of the CDB technique on the classifier weights.
We start by normalizing the classifier parameters $\boldsymbol{\omega} = \boldsymbol{\omega}_{R}^s / ||\boldsymbol{\omega}_{R}^s||_2$ across the feature dimension, before computing and plotting the similarity matrix $\mathbf{S} = \boldsymbol{\omega} \, \boldsymbol{\omega}^\intercal \in \mathbb{R}^{C\times C}$.
This matrix captures the %
similarity between each class embedding and all the others.
High off-diagonal activations indicate cross-talk between classes, leading to low-confidence predictions, since a %
single feature will activate multiple classes at the same time.
CDB minimizes class bias by pushing the self-similarity matrix closer to an identity matrix, reducing cross-class interference.
In the figure, we compare the matrices computed on the last aggregated models trained with the CDB block disabled (left) and with CDB turned on (right).
We can appreciate how disabling the block leads to significant cross-talk and to confused predictions, while the full FedPromo framework (thanks to CDB) substantially reduces bias, resulting in sharper, more confident classifications.

\begin{table}[t]
    \renewcommand{\arraystretch}{.9}
    \centering
    \begin{tabular}{c|c:c|c:c|c}
        \multirow{2}{*}{$\eta$} & \multicolumn{2}{c|}{MobileNet} & \multicolumn{2}{c|}{DINOv2} & \multirow{2}{*}{Avg.} \\
        & Top-1 & Top-5 & Top-1 & Top-5 \\
        \hline
        0.00 & 34.9 & 61.2 & 35.1 & 68.9 & 50.0 \\
        0.05 & 37.0 & 63.4 & 36.1 & 69.5 & 51.5 \\
        0.10 & 38.1 & 63.7 & 35.4 & 67.4 & 51.2 \\
        0.50 & 30.4 & 50.5 & 23.0 & 43.4 & 36.8 \\
    \end{tabular}
    \caption{Tuning of $\eta$. We focus on the best overall performance. Federated Top-1 and Top-5 accuracy ($\uparrow$) on  CompCars (431 classes)  starting from StanfordCars (196 classes).}
    \label{tab:eta}
\end{table}

\section{Conclusions} \label{sec:conclusions}
We introduced FedPromo, a novel federated learning framework to adapt a large server-side foundation model to local client data without sharing such data to preserve privacy and lower communication costs. 
Our approach pretrains a lightweight proxy model to generate feature representations closely aligned with the foundation model. 
The proxy model is used to train a classifier at remote clients in a federated setting, allowing seamless knowledge transfer. 
Once trained, the classifier is detached from the proxy model and attached to the foundation model, improving its generalization without direct access to client data.
Experimental results demonstrate the effectiveness of the proposed approach on multiple datasets.

Future research will explore alternative and flexible distillation strategies, enabling the extension of the approach to additional vision tasks such as object detection and semantic segmentation.

\section*{Acknowledgements}
This work was partially supported by the European Union under the Italian National Recovery and Resilience Plan (NRRP) of NextGenerationEU, partnership on "Telecommunications of the Future" (PE00000001- program "RESTART").

\newcommand{\disablebib}{}
\setcounter{section}{0}
\setcounter{figure}{0}
\setcounter{table}{0}
\setcounter{algorithm}{0}
\ifx\disablebib\undefined
\clearpage
\maketitlesupplementary
\else
{
    \small
    \bibliographystyle{ieeenat_fullname}
    \bibliography{main}
}
\fi

\renewcommand{\thesection}{S.\arabic{section}}
\renewcommand{\thefigure}{S.\arabic{figure}}
\renewcommand{\thetable}{S.\arabic{table}}
\renewcommand{\thealgorithm}{S.\arabic{algorithm}}

\ifx\disablebib\undefined
\noindent
This document presents some additional material that could not be included in the paper due to space constraints.
Specifically, we provide: 
in-depth information about the ImageNetPets dataset (Sec.~\ref{sec:supp:dts}); 
further visualizations of the alignment of the feature distribution (Sec.~\ref{sec:supp:alignment});
the per-round validation accuracy analysis (Sec.~\ref{sec:supp:evaluation}); 
additional details on the multi-domain performance (Sec.~\ref{sec:supp:multi-domain});
the performance using a smaller server model (Sec.~\ref{sec:supp:architecture});
the data distribution across clients used in our federated setup (Sec.~\ref{sec:supp:client});
some additional ablation analysis (Sec.~\ref{sec:sup:additional-ablation});
a detailed explanation of the $(\epsilon, \delta)$-DP parameters (Sec.~\ref{sec:supp:dp-parmas});
a statistical robustness analysis (Sec.~\ref{sec:supp:statistics});
a description of the implementation details (Sec.~\ref{sec:supp:implementation});
the pseudocode for the main modules of the FedPromo framework (Sec.~\ref{sec:supp:pseudocode});
the limitations of our work (Sec.~\ref{sec:supp:limits});
and ethical concerns related to our work (Sec.~\ref{sec:supp:ethical}).
\else
{%
\newpage
\centering
\LARGE
\textbf{Supplementary Material}\\[1.5ex]
}
\fi

\section{Additional details on ImageNetPets} \label{sec:supp:dts}
The OxfordPets dataset \cite{parkhi2012cats} contains images of $25$ dogs and $12$ cats breeds. To obtain a domain counterpart, we decided to extract samples from ImageNet-1k \cite{imagenet_cvpr09} since it already contains many images of cats and dogs with a more coarse breed classification. More precisely following the Wordnet \cite{christiane2005wordnet} hierarchy we kept all the dog and cat samples splitting them into their high level subcategories, achieving a total of $14$ dogs and $2$ cats classes as shown in Tab.~\ref{tab:supp:imnetpetsclasses}.

\section{Alignment of Feature Representations with Knowledge Distillation} \label{sec:supp:alignment}
\begin{figure*}[ht]
    \centering
    \includegraphics[width=0.8\linewidth]{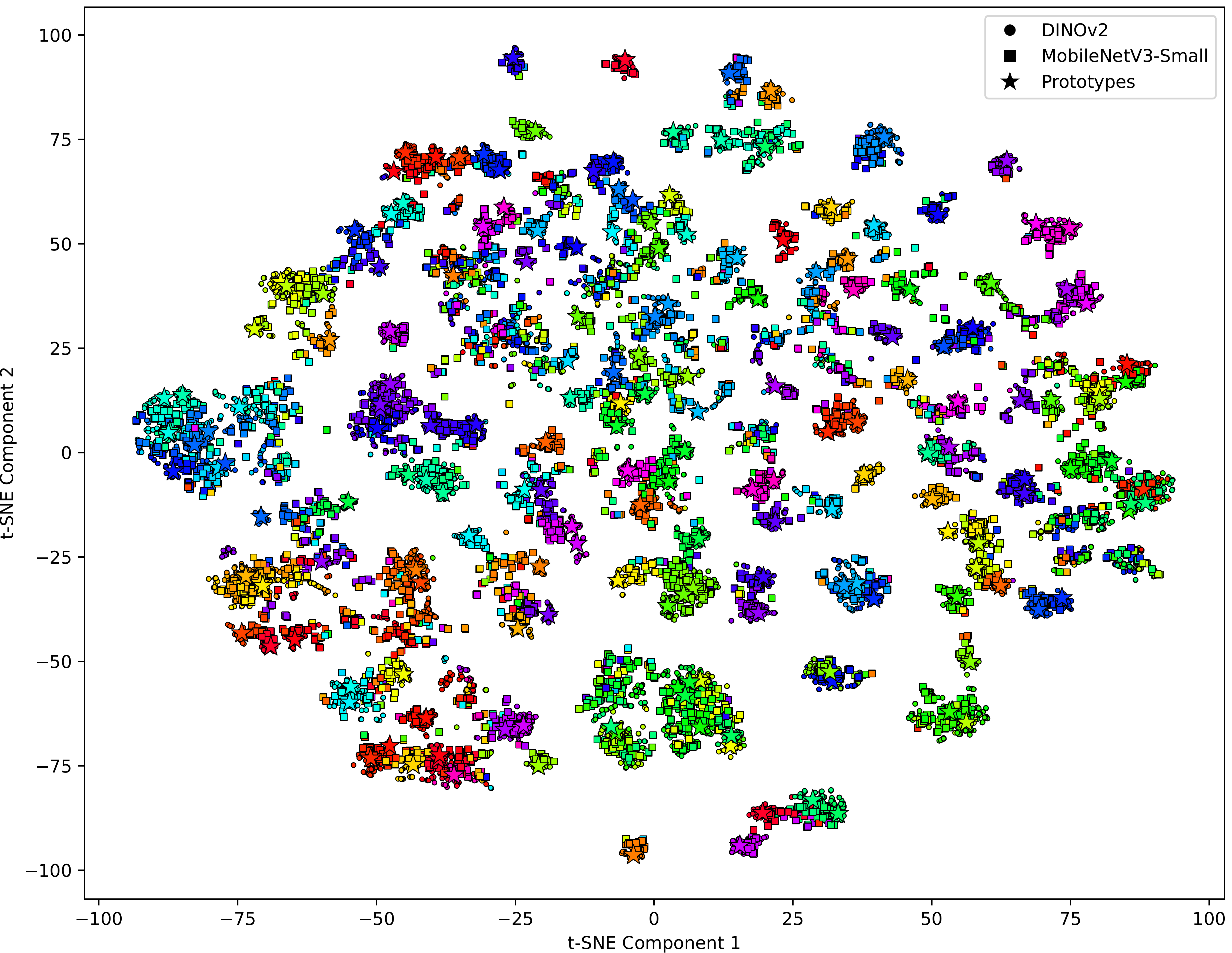}
    \caption{t-SNE plot showing the alignment rate of the features after pretraining on the StanfordCars dataset on all classes. Due to the class density multiple classes are mapped in the same color.}
    \label{fig:supp:tsne}
\end{figure*}

Fig. \ref{fig:supp:tsne} presents a detailed visualization of feature alignment achieved through our pretraining strategy. Unlike the reduced version in the main paper, this plot includes all classes, offering a denser but more informative plot. The alignment between DINO features ($\circ$) and MobileNet features ($\square$) remains consistent across classes, confirming the effectiveness of our approach.
Additionally, we report the DINO class prototypes ($\medwhitestar$), which highlight the intrinsic difficulty of the StanfordCars dataset. For example, some clusters (\eg, the leftmost one close to $y=0$) contain multiple prototypes, reflecting high inter-class similarity.

\begin{figure*}[ht]
    \centering
    \includegraphics[width=0.8\linewidth]{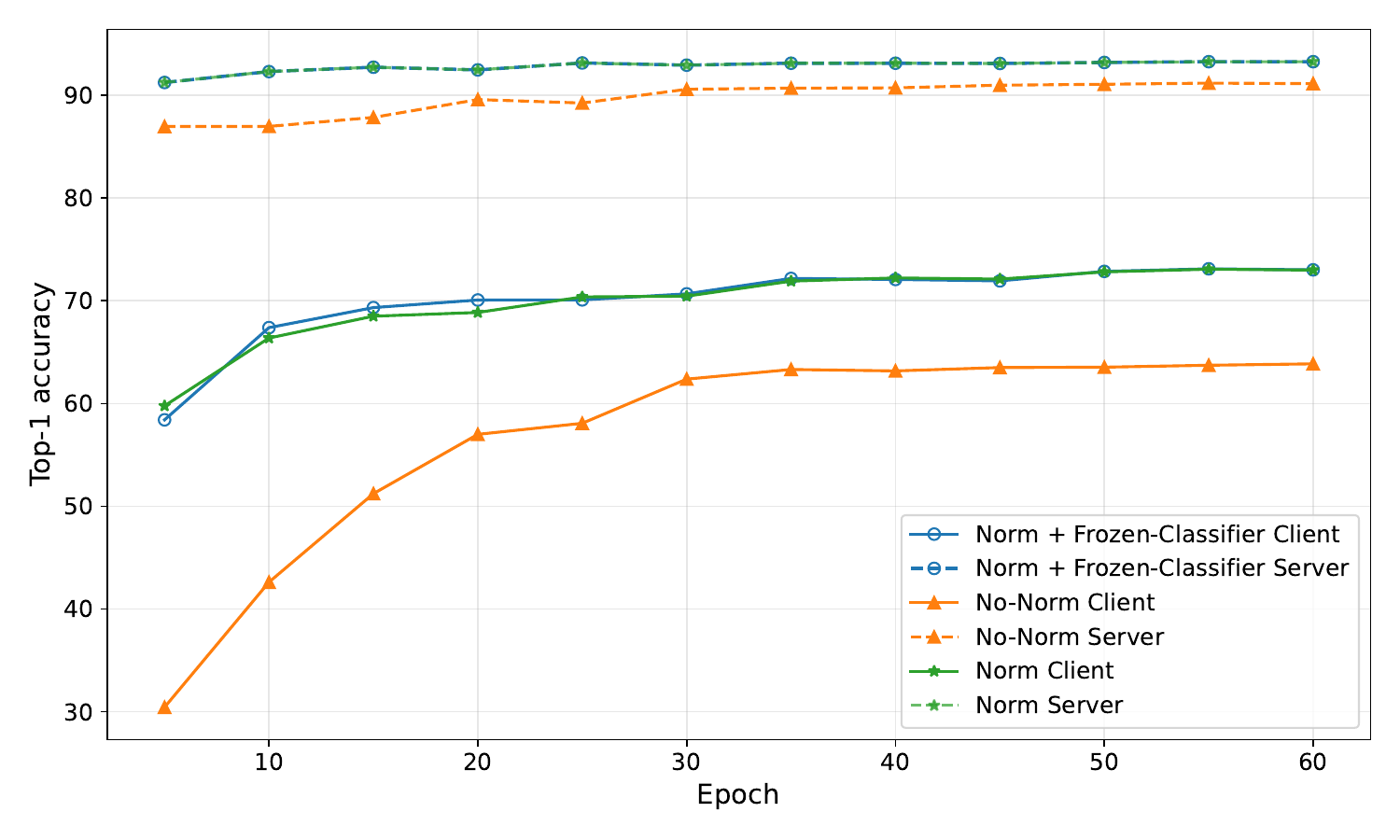}
    \caption{Effect of L2-normalization on the features during pretraining. Top-1 Accuracy on the StanfordCars dataset.}
    \label{fig:l2norm}
\end{figure*}

In Fig. \ref{fig:l2norm}, we show the accuracy gains obtained by incorporating L2-normalization at  feature level in pretraining.

We report three pairs of curves (\ie, the Top-1 accuracy at client and at server-side for each experiment) changing the pretraining architecture configuration.
\begin{itemize}
    \item In \textcolor{orange}{orange}, we show the validation accuracy when no normalization is applied to the client and server features, highlighting how this leads to lackluster results.
    \item In \textcolor{green}{green}, we show a similar curve, but we enable our L2-normalization layer on both server and client models, leading to an impressive performance gain.
    \item In \textcolor{blue}{blue}, we report the accuracy curve attainable when training the classifier only on server features and keeping it frozen when applied to the client features. This leads to a further small improvement in accuracy.
\end{itemize}

The data in the plot confirms that normalizing feature magnitudes improves alignment between server and client encoder outputs and contributes to the final performance of the federated models.
We remark that the server accuracies of \textit{Norm + Frozen-Classifier} and \textit{Norm} are overlapping because, from the server model (DINO) point of view, nothing is changing. The \textit{Frozen-Classifier} mark refers to MobileNet's guidance, which is disabled; gradient through the  DINO encoder is still enabled.

\begin{table}[t]
    \centering
    \begin{tabular}{c|cc}
        Animal &  \multicolumn{2}{c}{Classes} \\
        \hline
        Cats & Domestic cat & Wild cat \\
        \hdashline
        \multirow{7}{*}{Dogs} & Hunting dog & Toy dog \\
        & Working dog & Corgi \\
        & Griffon & Poodle \\
        & Spitz & Basenji \\
        & Newfoundland & Mexican hairless \\
        & Leonberg & Great Pyrenees \\
        & Dalmatian & Pug \\
    \end{tabular}
    \caption{ImagenetPets classes}
    \label{tab:supp:imnetpetsclasses}
\end{table}

\section{Per-round Validation Accuracy} \label{sec:supp:evaluation}
\begin{figure*}
    \centering
    \begin{subfigure}{\linewidth}
        \centering
        \includegraphics[width=0.8\linewidth]{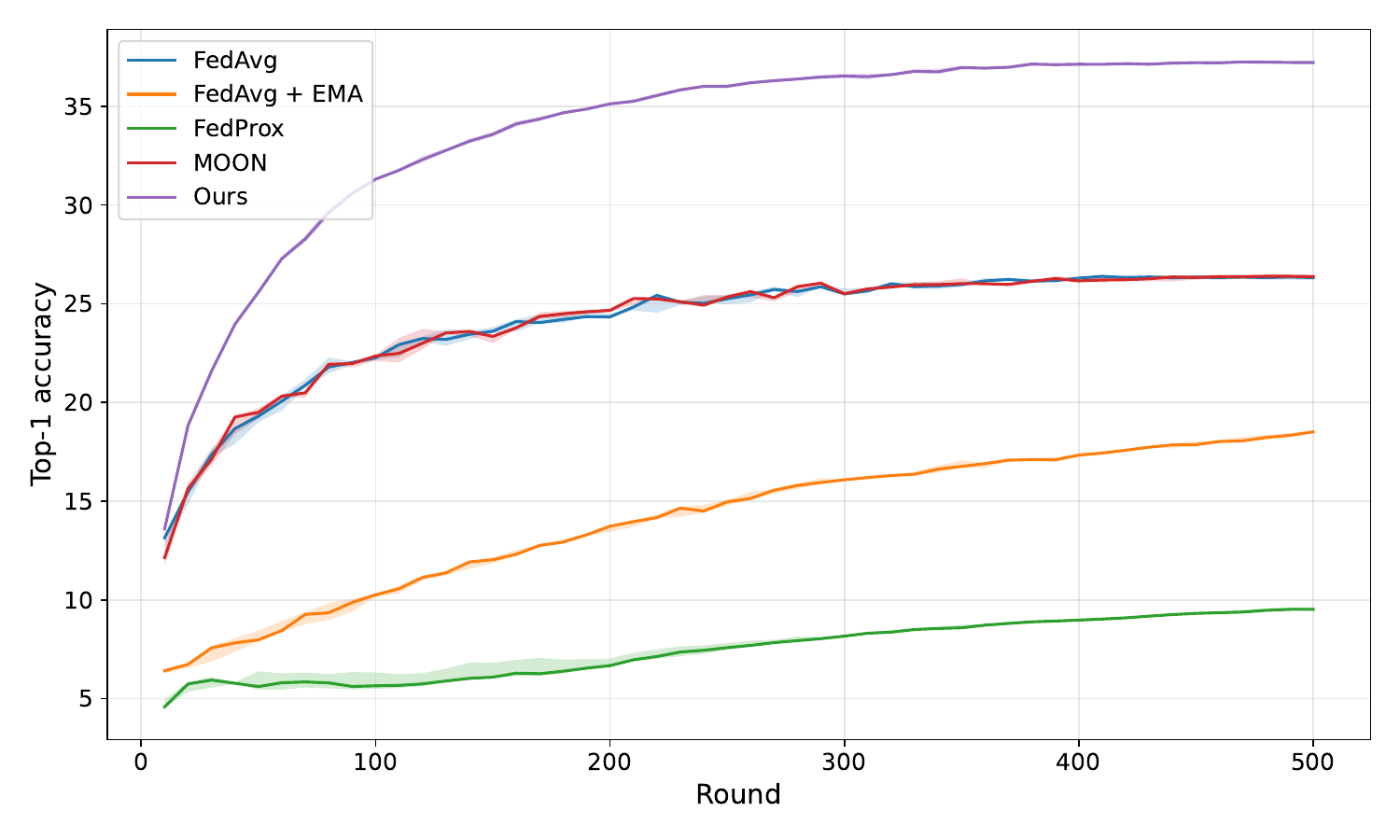}
        \caption{MobileNet}
        \label{fig:evaluation-curves:mobilenet}
    \end{subfigure}

    \vspace{0.5cm}

    \begin{subfigure}{\linewidth}
        \centering
        \includegraphics[width=0.8\linewidth]{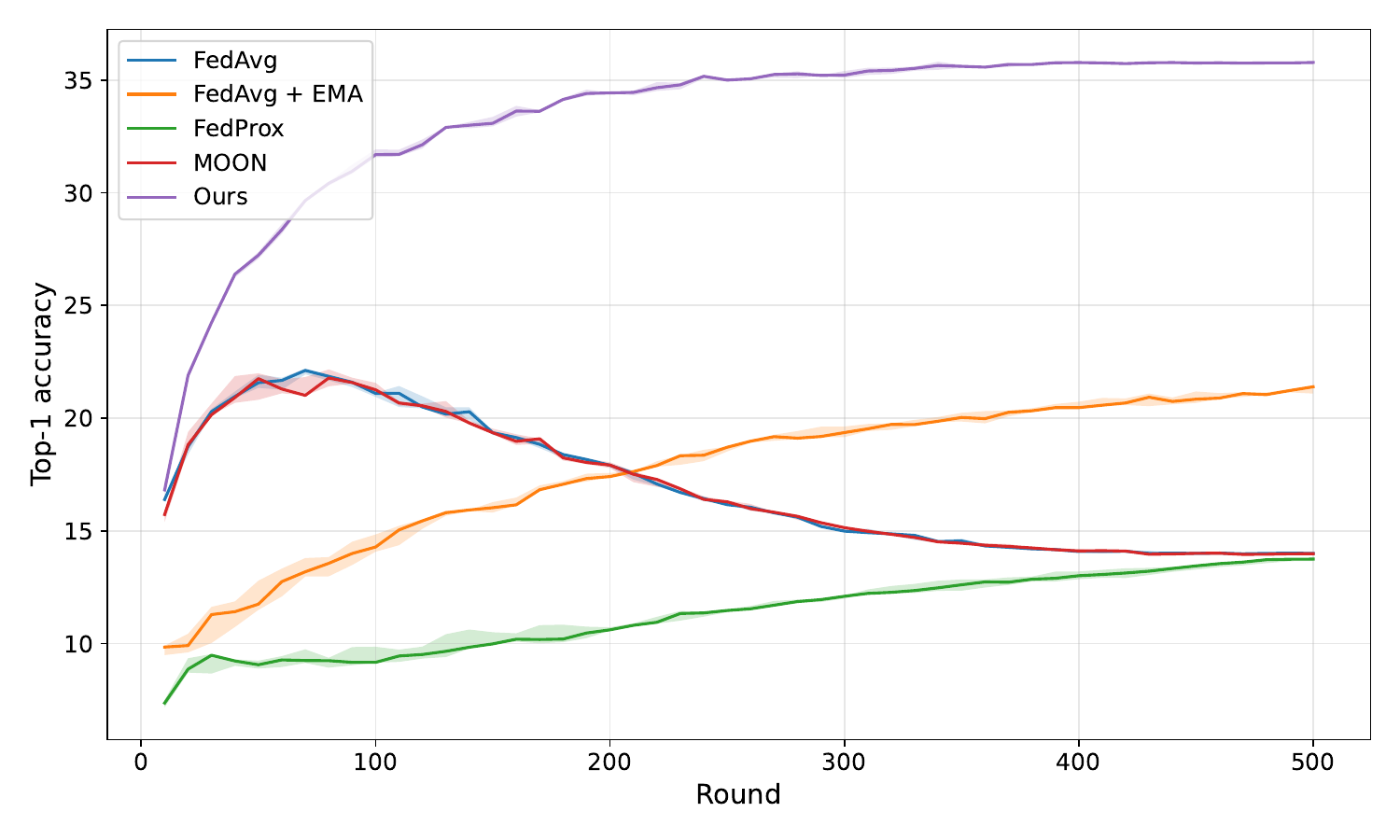}
        \caption{DINOv2}
        \label{fig:evaluation-curves:dinov2}
    \end{subfigure}

    \caption{Evaluation curves of the Top-1 accuracy with confidence intervals (shaded) between first and third quartile on the CompCars dataset during the Federated Learning rounds.}
    \label{fig:evaluation-curves}
\end{figure*}

Fig.~\ref{fig:evaluation-curves} displays the per-round accuracy curves with confidence intervals (shaded) between first and third quartiles on the validation set of the CompCars dataset across different approaches.
These curves demonstrate that our proposed method: (i) achieves higher accuracy compared to competitors; (ii) converges faster, reaching peak performance in fewer rounds; (iii) maintains stability throughout training, avoiding oscillations or performance drops.

This visualization further reinforces the effectiveness of our approach in federated learning settings.

\section{Multi-Domain Evalution}
\label{sec:supp:multi-domain}
\begin{figure*}[htbp]
    \centering
    \includegraphics[width=0.965\linewidth]{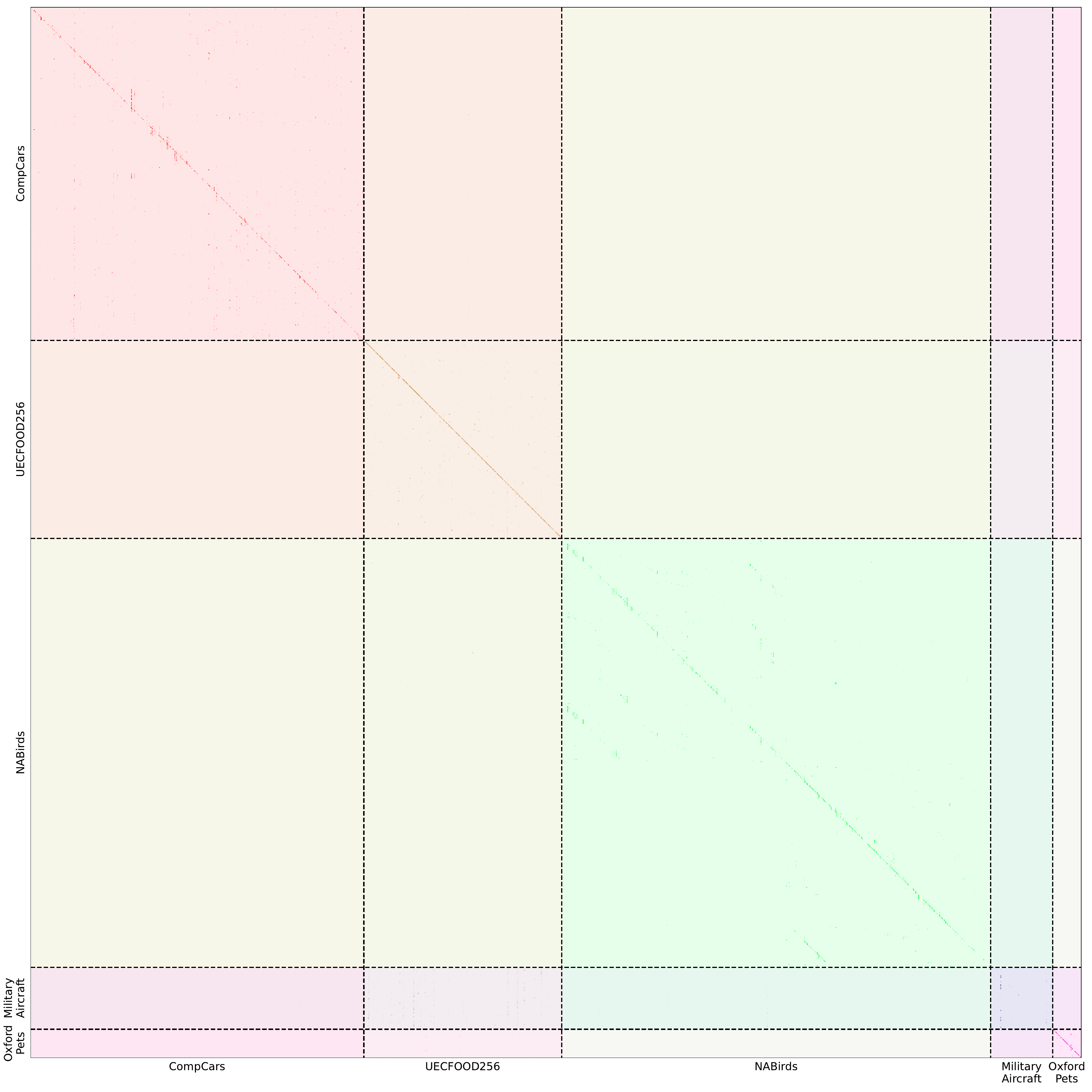}
    \caption{Multi-domain confusion matrix}
    \label{fig:multi_domain_cf}
\end{figure*}

\begin{figure*}[htbp]
    \centering
    \includegraphics[width=0.965\linewidth]{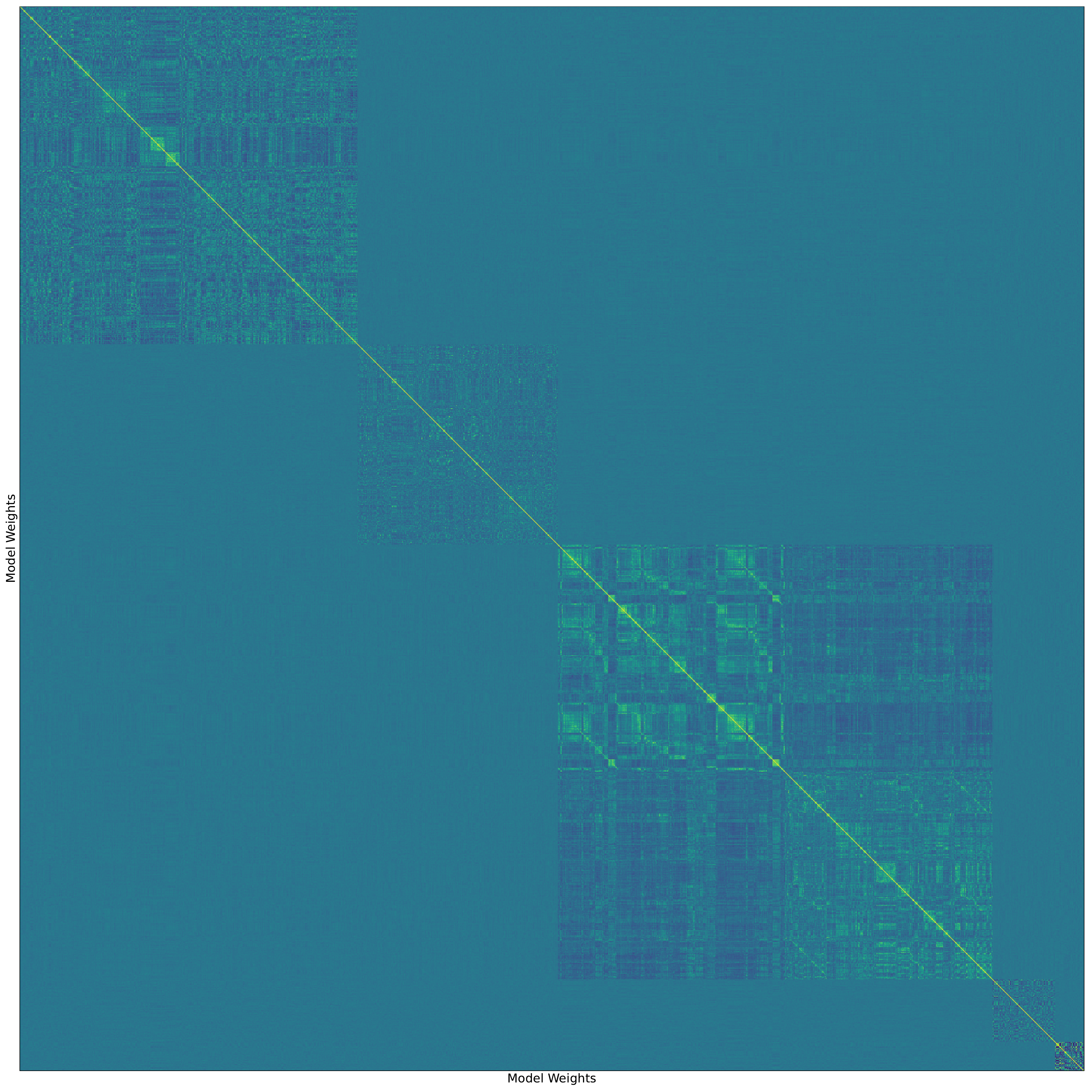}
    \caption{Multi-domain classifier weights self-similarity}
    \label{fig:multi_domain_self-sim}
\end{figure*}

Fig.~\ref{fig:multi_domain_cf} represents the confusion matrix of the multi-domain classification test using all the classifiers trained during the different federated learning trainings concatenated together into a single classifier. Fig.~\ref{fig:multi_domain_self-sim} visualizes the self-similarity of the aggregated classifier weights.
These two images show that, also in the multi-domain test, the classifier has neglegible cross-talk between classes belonging to different datasets. Only the Military Aircraft dataset suffers slightly from this phenomenon due to the low performance of the model on this data during training.

\section{Server Model Ablation: DINOv2-Small}
\label{sec:supp:architecture}
To confirm the adaptability of FedPromo to other deep networks, as well as to study the performance change with respect to the capabilities of the server model, in this section, we report a comparison between FedPromo and FedAvg when we use  DINOv2-small as the server model.
The experiments on CompCars confirm the findings of the main paper, as FedPromo outperforms FedAvg by more than $18\%$ even in this scenario. The values also highlight the necessity of accurate choice of the server model, as using a smaller network decreases the average accuracy from $51.5\%$ to $42.3\%$.

\begin{table}[bht]
    \centering
    \begin{tabular}{c:c|c:c}
        $\mathcal{D}^s$ & $C$ & $\mathcal{D}$ & $C$ \\
        \hline
        StanfordCars~\cite{krause20133d} & $196$ & CompCars~\cite{yang2015large} & $431$ \\
        Food101~\cite{bossard2014food} & $101$ & UECFOOD256~\cite{kawano14c} & $256$ \\
        CUB200~\cite{wah2011caltech} & $200$ & NaBirds~\cite{Horn_2015_CVPR} & $555$ \\ %
        FGVCAircraft~\cite{maji13fine-grained} & $100$ & MilitaryAircraft~\cite{militaryAricraft} & $80$ \\
        ImageNetPets & $16$ & OxfordPets~\cite{parkhi2012cats} & $37$ \\
    \end{tabular}
    \caption{Number of classes in each pair of pretraining ($\mathcal{D}^s$) and fine-grained target ($\mathcal{D}$) datasets.}
    \label{tab:class_count}
\end{table}

\section{Data Distribution}
\label{sec:supp:client}
\begin{figure*}[htbp]
    \centering
    \includegraphics[height=0.965\textheight]{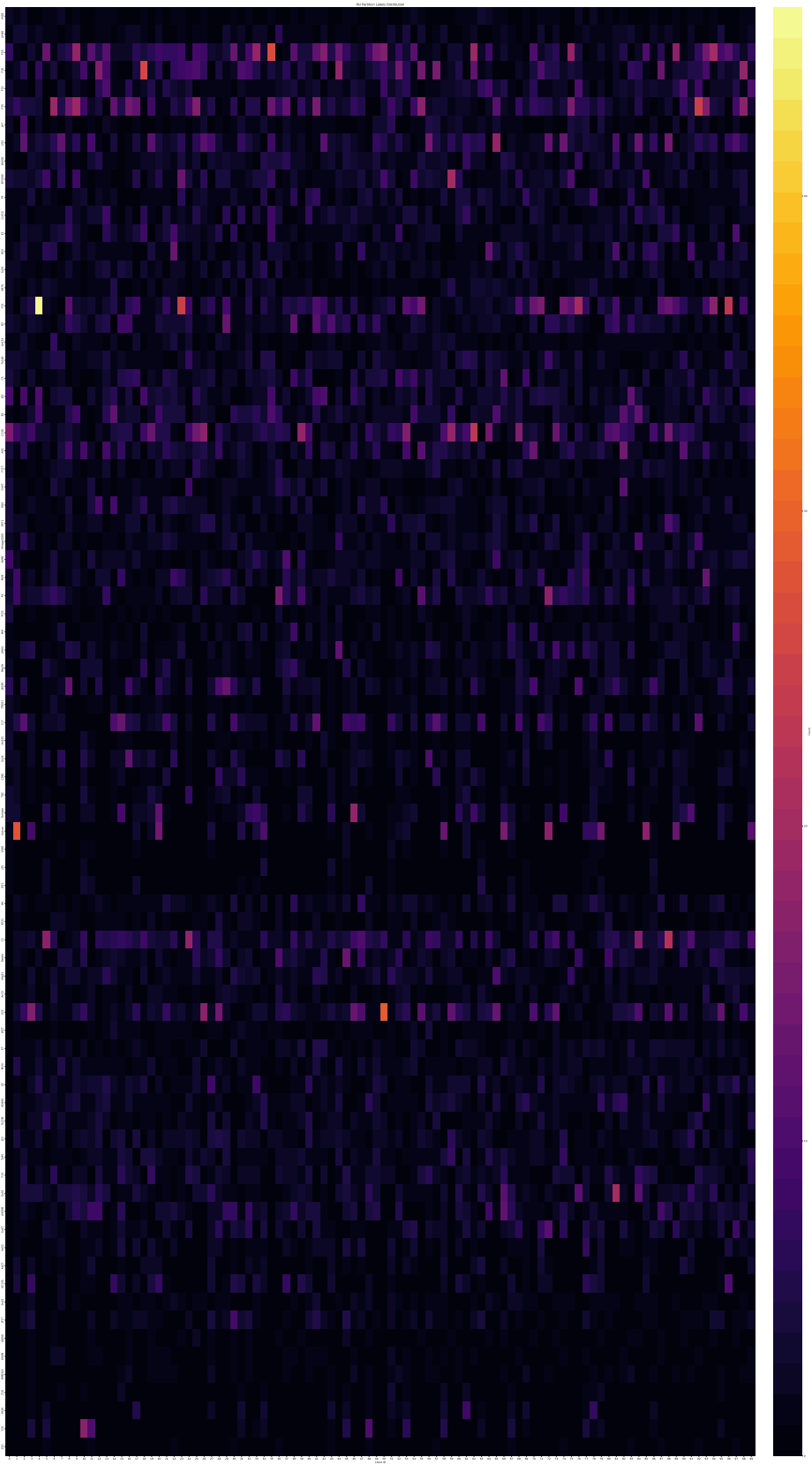}
    \caption{Per client class distribution on the MilitaryAircraft \cite{militaryAricraft} dataset.}
    \label{fig:militaryaircraft_heatmap}
\end{figure*}

Figs.~\ref{fig:compcars_heatmap}-\ref{fig:oxfordpets_heatmap} illustrate the data distribution across clients in the federated learning setup.
Each row represents a class, and each column corresponds to a client.
The color intensity indicates the number of samples of each class assigned to each client.
These heatmaps provide insight into the heterogeneity of the data distribution and the challenges posed by non-IID splits in federated learning.
In Tab.~\ref{tab:class_count} the number of classes contained in all the considered datasets is also shown for reference.

\section{Additional Ablation} \label{sec:sup:additional-ablation}
To confirm the findings attained in the ablation performed in the main paper, in Tab.~\ref{tab:in_domain_cdb} we show the comparison between enabling or disabling CDB on two additional datasets (i.e., UECFOOD256 and NABirds). As before, we find a small increase in performance with a minimal computational footprint, as well as the weight-regularization effect.

\begin{table}[hbt]
    \centering
    \small
    \begin{tabular}{c|c|c:c|c:c|c}
        \multirow{2}{*}{Dset%
        } & \multirow{2}{*}{CDB} & \multicolumn{2}{c|}{MobileNet} & \multicolumn{2}{c|}{DINOv2} & \multirow{2}{*}{Avg.} \\
        & & Top-1 & Top-5 & Top-1 & Top-5 \\
        \hline
        \multirow{2}{*}{\includegraphics[width=16pt]{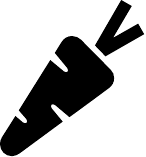}} & \cmark & 52.1 & 80.0 & 58.7 & 88.2 & 69.8 \\
                                    & \xmark & 51.9 & 80.0 & 58.7 & 88.2 & 69.7 \\
        \hdashline
        \multirow{2}{*}{\includegraphics[width=16pt]{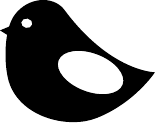}} & \cmark & 32.2 & 59.2 & 30.3 & 65.0 & 46.7 \\
                                 & \xmark & 32.2 & 59.0 & 30.3 & 65.0 & 46.6 \\
        \hline
        \multirow{2}{*}{Avg.} & \cmark & 42.2 & 69.6 & 44.5 & 76.6 & 58.3 \\
                              & \xmark & 42.1 & 69.5 & 44.5 & 76.6 & 58.2 \\
    \end{tabular}
    \caption{Analysis on the impact of CDB in in-domain datasets (\ie, UECFOOD256 \includegraphics[width=10pt]{icons/food_icon.pdf} and NABirds \includegraphics[width=10pt]{icons/bird.pdf}).}
    \label{tab:in_domain_cdb}
\end{table}

\section{Differential Privacy Parameters} \label{sec:supp:dp-parmas}
In order to properly reach $(\epsilon, \delta)$-DP, we need to set some other parameters in addition to  $\epsilon$ and $\delta$.

\begin{figure}[ht]
    \centering
    \includegraphics[width=1\linewidth]{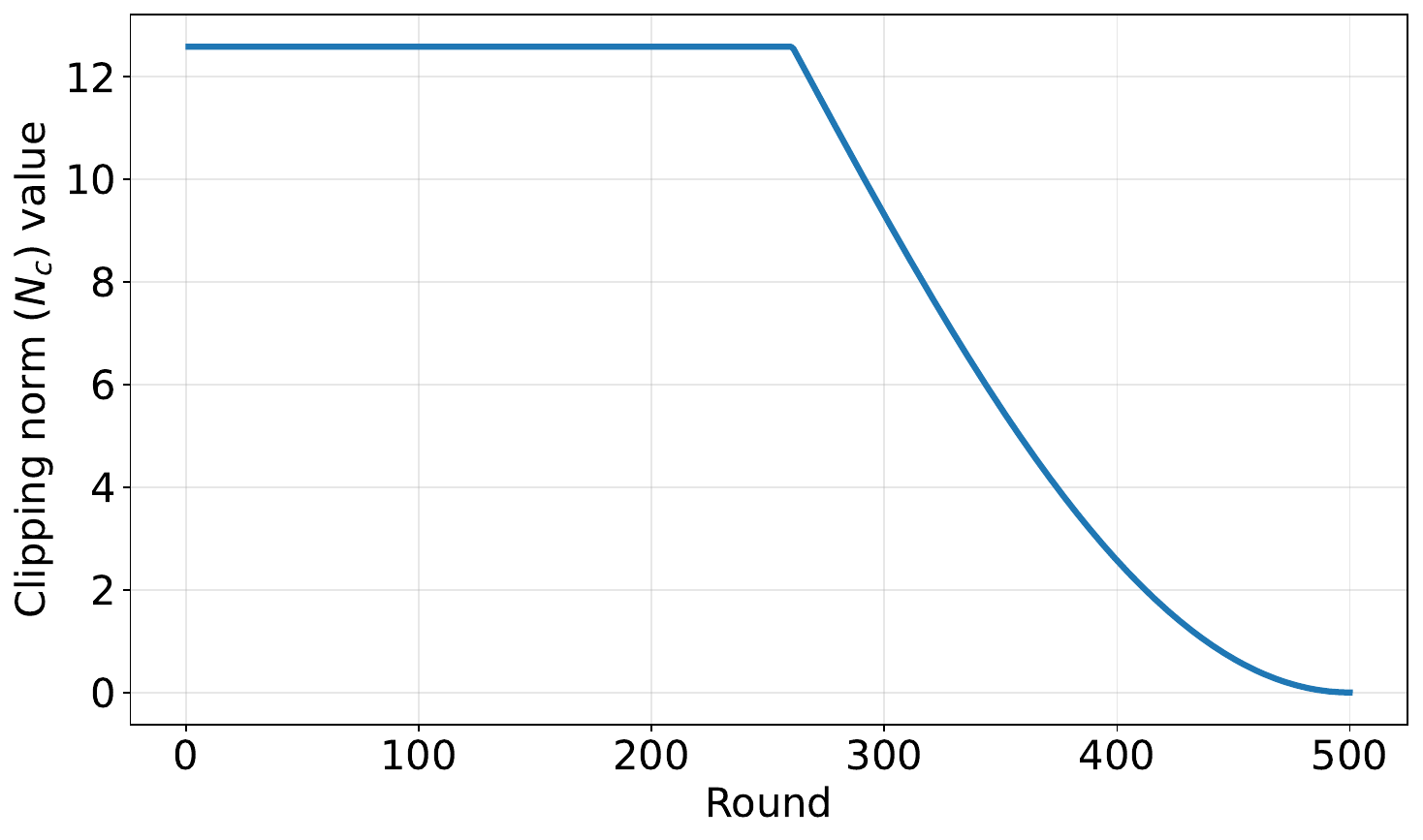}
    \caption{Clipping norm scheduler for the CompCars dataset}
    \label{fig:clipping_norm}
\end{figure}

\noindent \textbf{Clipping Norm ($\boldsymbol{N_c}$):}
According to \cite{abadi2016deep}, the privacy budget can be decoupled from the strength of the injected Gaussian noise before transmission if a sufficiently low clipping norm $N_c$ is applied to the gradients before adding the noise (which is used to compute an appropriate noise scale).
Since we employ learning rate scheduling, which has the side effect of reducing gradient norms over time, we had to adapt the clipping norm $N_c$ accordingly.
\cite{wei2020federated} argues that a good choice of norm is the median of norms over the full evolution of training. Following their insight, we set the maximum clipping norm at  $\hat{N}_c = 12.6$ (for CompCars) and change its value over rounds following the learning rate scheduler:
\begin{equation}
    N_c[r] = \min\left(\hat{N}_c, N_{c,\text{max}}\times \cos\left(\frac{\pi}{2 R} r \right)^2\right)
\end{equation}
where $N_{c,\text{max}} = 26.9$,  $r$ is the current round and $R$ the total number of rounds. In Fig. \ref{fig:clipping_norm} it is possible to see the $N_c$ value for each federated round on the CompCars datasets.

\noindent \textbf{Sensitivity ($\boldsymbol{\Delta_s}$):}
The sensitivity parameter controls the strength of the noise injected before transmission. Following \cite{wei2020federated} we define it as a function of $N_c$ as follows (for the CompCars dataset):
\begin{equation}
    \Delta_s = 2 \cdot \frac{N_c}{\min_k\{|\mathcal{D}_k|\}} = \frac{N_c}{65}
\end{equation}

\begin{table}[t]
    \centering
    \begin{tabular}{c|c:c}
        Compared Method & Top-1 & Top-5 \\
        \hline
        FedAvg~\cite{mcmahan2017communication} & $2 \times 10^{-5}$ & $7 \times 10^{-5}$ \\
        FedAvg~\cite{mcmahan2017communication} + EMA & $2 \times 10^{-5}$ & $8 \times 10^{-5}$ \\
        FedProx~\cite{li2020federated} & $3 \times 10^{-6}$ & $7 \times 10^{-6}$ \\
        MOON~\cite{li2021model} & $2 \times 10^{-5}$ & $1 \times 10^{-6}$ \\
    \end{tabular}
    \caption{$p$-value statistics of the Wilcoxon signed rank metric comparing FedPromo to its competitors.}
    \label{tab:wilcoxon}
\end{table}

\begin{table}[t]
    \centering
    \begin{tabular}{c|c:c|c:c}
        \multirow{2}{*}{Method} & \multicolumn{2}{c|}{Client} & \multicolumn{2}{c}{Server} \\
        & last & avg & last & avg \\
        \hline
        FedAvg~\cite{mcmahan2017communication} & $0.2\%$ & $0.3\%$ & $0.1\%$ & $0.2\%$ \\
        FedAvg~\cite{mcmahan2017communication} + EMA & $0.2\%$ & $0.3\%$ & $0.2\%$ & $0.4\%$ \\
        FedProx~\cite{li2020federated} & $0.1\%$ & $0.2\%$ & $0.1\%$ & $0.3\%$ \\
        MOON~\cite{li2021model} & $0.1\%$ & $0.3\%$ & $0.1\%$ & $0.3\%$ \\
        FedPromo (ours) & $0.1\%$ & $0.1\%$ & $0.1\%$ & $0.2\%$ \\
    \end{tabular}
    \caption{Last round and average standard deviations of Top-1 Accuracy on CompCars dataset. For each model, we consider $5$ different runs varying %
    the RNG seed.}
    \label{tab:std}
\end{table}

\section{Statistical Robustness Analysis}
\label{sec:supp:statistics}
To verify the statistical significance of the results in the experimental evaluation, we have computed the Wilcoxon signed rank metric for the experiments reported in Tab.~1 and Tab.~2.
We compute the metric on all available pairs of accuracies of the same type (top-1/top-5).
In the computation, we include client and server accuracies, as well as in-domain and out-of-domain scores.
We report the null hypothesis p-value of these studies in Tab.~\ref{tab:wilcoxon}.
To increase the robustness of our statistical analysis, we also performed the Friedman test on the same data, reaching an overall \mbox{p-value} of $2 \times 10^{-6}$ in Top-1 and $3 \times 10^{-7}$ in Top-5.
We achieve $p < 0.001$ in all cases, confirming the strength of FedPromo.

Finally, in Tab.~\ref{tab:std} we report the last round and average standard deviations of the Top-1 accuracy on the CompCars dataset varying the Random Number Generator (RNG) seed (21, default=\textbf{42}, 98, 198, 2025). Our model proves to  consistently reach a standard deviation under $0.2\%$.

\section{Further Implementation Details}
\label{sec:supp:implementation}
We run all of our experiments on a RHEL8 (RedHat) Unix-based machine with kernel version 6.12.9-1, equipped with 8 NVIDIA L40S GPUs (48GB of VRAM, CUDA 12.8, Driver 570.86.15), 2 AMD EPYC 9224 24-Core processors, 1.5TB of RAM, and Python version 3.11.9. Each federated training experiment uses a single GPU, 6 CPU cores, and 32GB of RAM. In the CompCars dataset, a complete training pipeline takes about 9 hours on a single GPU.
Fine-grained information on all packages used in the environment is available as an \texttt{environment.yml} file in the code repository.

\section{FedPromo Pseudocode}
\label{sec:supp:pseudocode}
In this section, we report the pseudocode implementation of FedPromo.
Algorithm~\ref{alg:icp} reports the steps necessary to perform ICP, while Algorithm~\ref{alg:cdb} explains how CDB is applied to the model weights.
Algorithm~\ref{alg:server_round} describes the steps taken by the server to coordinate and aggregate client updates. Algorithm~\ref{alg:client_round} details the local optimization performed by each client.

\renewcommand{\algorithmicensure}{\textbf{Input:}}

\begin{algorithm}[b]
\caption{Pseudocode for FedPromo's ICP.}
\label{alg:icp}
\begin{algorithmic}[1]
\REQUIRE ICP hyperarameter $\eta$, class set $\mathcal{C}$
\ENSURE Current Decoder $D_e$, previous epoch's decoder $D_{e-1}$, active classes set $\mathcal{A}$
\STATE Extract the decoders' parameters $\boldsymbol{\omega}_e \in \mathbb{R}^{C \times F}$ from $D$, and $\boldsymbol{\omega}_{e-1} \in \mathbb{R}^{C \times F}$ from $D_{e-1}$
\FOR{$c \in \mathcal{C} \setminus \mathcal{A}$}
    \STATE $\boldsymbol{\omega}_e[c] \gets \eta \boldsymbol{\omega}_e[c] + (1-\eta) \boldsymbol{\omega}_{e-1
    }[c]$
\ENDFOR
\STATE Update $D_e$ with the new parameters $\boldsymbol{\omega}_e$
\RETURN $D_e$
\end{algorithmic}
\end{algorithm}

\begin{algorithm}[b]
\caption{Pseudocode for FedPromo's CDB.}
\label{alg:cdb}
\begin{algorithmic}[1]
\ENSURE Decoder $D$ with parameters $\boldsymbol{\omega} \in \mathbb{R}^{C \times F}$.
\STATE Average the entries of $\boldsymbol{\omega}$ over the class dimension: $\bar{\boldsymbol{\omega}}[\mathrm{f}] \gets \frac{1}{C}\sum_{c=1}^{C} \boldsymbol{\omega}[\mathrm{f}], \;\; \forall\mathrm{f} = 1\dots F$
\STATE Remove the bias component from the parameters $\boldsymbol{\omega}$: $\boldsymbol{\omega}[\mathrm{f}] \gets \boldsymbol{\omega}[\mathrm{f}] - \bar{\boldsymbol{\omega}}[\mathrm{f}], \;\; \forall\mathrm{f} = 1\dots F$
\STATE Update $D$ with the new parameters $\boldsymbol{\omega}$
\RETURN $D$
\end{algorithmic}
\end{algorithm}

\begin{algorithm}
\caption{Pseudocode for FedPromo ServerRound.
For simplicity, we report the code for a single domain $\mathcal{D}^i$.
}
\label{alg:server_round}
\begin{algorithmic}[1]
\REQUIRE Number of rounds $R$ and active clients $K_A$, domain index $i$, server encoder $O$, peak learning rate $lr_{\text{max}}$, cosine annealing learning rate scheduler (see Sec. 5)
\STATE Select the appropriate clients group $\mathcal{K}^i$, client encoder $E$ and shared decoder $D$ for the domain
\STATE Send to all clients $c \in \mathcal{K}^i$ the encoder $E$ and decoder $D$
\FOR {$r \gets 1\dots R$}
    \STATE Compute the active clients set $\mathcal{K}_A$ by randomly sampling $K_A$ clients from $\mathcal{K}^i$ \COMMENT{Different on each round}
    \STATE Compute current round learning rate from the scheduler $lr \gets \text{CosineSchedule}(lr_{\text{max}}, r, R)$
    \FOR[Performed in Parallel]{$k \in \mathcal{K}_A$}
        \STATE Receive $D^k \gets \text{ClientRound}(D, r, lr)$ from $k$
    \ENDFOR
    \STATE Aggregate decoders $D \gets \text{FedAvg}(\left\{D^k\right\}_{k \in \mathcal{K}_A})$
    \STATE Debias decoder $D \gets \text{CDB}(D)$
\ENDFOR
\end{algorithmic}
\end{algorithm}

\begin{algorithm}
\caption{Pseudocode for FedPromo ClientRound.}
\label{alg:client_round}
\begin{algorithmic}[1]
\REQUIRE Encoder $E$, total number of local epochs $e_{\text{max}}$, client index $k$, batch size $B$
\ENSURE Server's previous round's decoder $D_{r-1}$, round number $r$, and current $lr$
\STATE Initialize local model $M^k \gets D_{r-1} \circ E$ %
\STATE Define $D_{e=0}^k \gets D_{r-1}$
\FOR {$e \gets 1\dots e_{\text{max}}$}
    \FOR {$b \gets 1\dots \lfloor |\mathcal{D}| / B \rfloor$}
        \STATE Extract $B$ samples from local dataset $\mathcal{D}$ into $\mathcal{B}$
        \STATE Initialize batch loss $l \gets 0$
        \STATE Initialize active classes set $\mathcal{A} \gets \varnothing$
        \FOR {$(\mathbf{X}, y) \in \mathcal{B}$}
            \STATE Update $\mathcal{A} \gets \mathcal{A} \cup \{y\}$
            \STATE $\mathbf{p} \gets M^k(\mathbf{X})$ \COMMENT{Model predictions}
            \STATE $l \gets l + \mathcal{L}_{CE}(\mathbf{p}, y)$ \COMMENT{Accumulate loss}
        \ENDFOR
        \STATE Update model $M^k \gets \text{AdamOptim}(M^k, l)$
    \ENDFOR
    \STATE Extract current model decoder $D^k_e$ from $M^k$
    \STATE Apply ICP: $D^k_{e} \gets \text{ICP}(D^k_e, D^k_{e-1}, \mathcal{A})$
    \STATE Update model $M^k \gets D^k_{e} \circ E$
\ENDFOR
\STATE Debias decoder $D^k_{r} \gets \text{CDB}(D^k_{e})$
\STATE Send $D^k_{r}$ to server
\end{algorithmic}
\end{algorithm}

\section{Limitations}
\label{sec:supp:limits}

FedPromo enables efficient and privacy-aware adaptation of foundation models in federated settings.
The method relies on access to an in-domain %
public dataset %
to achieve the best performance%
, which may not always be available. Performance can vary between domains, with weaker results in highly heterogeneous or challenging scenarios (\eg, MilitaryAircraft dataset). FedPromo is currently limited to classification tasks, and additional work is needed to extend it to more complex settings such as object detection or segmentation.

\section{Ethical Concerns}
\label{sec:supp:ethical}
While raw data remains on client devices, shared model components may still leak sensitive information %
if subject to attacks, highlighting the need for privacy-preserving and class-regularization techniques. Furthermore, decentralized learning may learn some local biases, potentially causing unfair outcomes if not addressed. Although current evaluations avoid these problems, caution is essential when handling sensitive private data such as facial images.

\FloatBarrier

\begin{figure*}[htbp]
    \centering
    \includegraphics[height=0.965\textheight]{suppl/compcars_heatmap.pdf}
    \caption{Per client class distribution on the CompCars \cite{yang2015large} dataset.}
    \label{fig:compcars_heatmap}
\end{figure*}

\begin{figure*}[htbp]
    \centering
    \includegraphics[height=0.965\textheight]{suppl/uecfood256_heatmap.pdf}
    \caption{Per client class distribution on the UECFOOD256 \cite{kawano14c} dataset.}
    \label{fig:uecfood256_heatmap}
\end{figure*}

\begin{figure*}[htbp]
    \centering
    \includegraphics[height=0.965\textheight]{suppl/nabirds_heatmap.pdf}
    \caption{Per client class distribution on the NABirds \cite{Horn_2015_CVPR} dataset.}
    \label{fig:nabirds_heatmap}
\end{figure*}

\begin{figure*}[htbp]
    \centering
    \includegraphics[height=0.965\textheight]{suppl/oxfordpets_heatmap.pdf}
    \caption{Per client class distribution on the OxfordPets \cite{parkhi2012cats} dataset.}
    \label{fig:oxfordpets_heatmap}
\end{figure*}

\end{document}